\documentclass[acmlarge]{acmart}

\makeatletter
\newcommand{\myconfshort}{\acmConference@shortname}
\newcommand{\myconffull}{\acmConference@name}
\newcommand{\myconfdate}{\acmConference@date}
\newcommand{\myconfloc}{\acmConference@venue}
\AtBeginDocument{
  \fancypagestyle{firstpagestyle}{
    \fancyhead{}%
    \fancyfoot[C]{}%
  }
  \fancyhf{}
  \fancyhead[LO]{\@headfootfont\shorttitle}%
  \fancyhead[RE]{\@headfootfont\@shortauthors}%
  \fancyhead[LE]{\@headfootfont\footnotesize \myconfshort, \myconfdate, \myconfloc}%
  \fancyhead[RO]{\@headfootfont\footnotesize \myconfshort, \myconfdate, \myconfloc}%
  \fancyfoot[C]{}%
}
\makeatother
\acmBooktitle{\conffull\@ (\confshort), \confdate, \confloc}

\usepackage{tcolorbox}
\tcbuselibrary{listingsutf8}
\usepackage{tabularx}
\usepackage{booktabs}
\usepackage{tikz}
\usetikzlibrary{arrows.meta,positioning,calc,fit}
\usepackage{enumitem}
\newlist{titemize}{itemize}{1}
\setlist[titemize]{leftmargin=1.2em,itemsep=0.15em,topsep=0pt,parsep=0pt,partopsep=0pt}

\AtBeginDocument{%
  }

\copyrightyear{2026}
\acmYear{2026}
\setcopyright{cc}
\setcctype{by}
\acmConference[FAccT '26]{The 2026 ACM Conference on Fairness, Accountability, and Transparency}{June 25--28, 2026}{Montreal, QC, Canada}
\acmBooktitle{The 2026 ACM Conference on Fairness, Accountability, and Transparency (FAccT '26), June 25--28, 2026, Montreal, QC, Canada}
\acmDOI{10.1145/3805689.3812324}
\acmISBN{979-8-4007-2596-8/2026/06}

\begin{document}

\title[A Scoping Review of the Ethical Perspectives on Anthropomorphising LLM-Based CAs]{A Scoping Review of the Ethical Perspectives on Anthropomorphising Large Language Model-Based Conversational Agents}

\author{Andrea Ferrario}
\authornote{Corresponding author.}
\orcid{0000-0001-9968-9474}
\email{aferrario@ethz.ch}

\affiliation{%
  \institution{Institute of Biomedical Ethics and History of Medicine, University of Z\"urich}
  \city{Z\"urich}
  \country{Switzerland}
}

\affiliation{%
  \institution{Dalle Molle Institute for Artificial Intelligence (IDSIA), SUPSI}
  \city{Lugano}
  \country{Switzerland}
}

\affiliation{%
  \institution{ETH Z\"urich}
  \city{Z\"urich}
  \country{Switzerland}
}

\author{Rasita Vinay}
\orcid{0000-0002-0490-5697}
\email{rasita.vinay@ibme.uzh.ch}

\affiliation{%
  \institution{Institute of Biomedical Ethics and History of Medicine, University of Z\"urich}
  \city{Z\"urich}
  \country{Switzerland}
}

 \affiliation{%
   \institution{Institute for Implementation Science in Health Care, University of Z\"urich}
   \city{Z\"urich}
   \country{Switzerland}
 }

\affiliation{%
  \institution{Department of Management, Technology and Economics, ETH Z\"urich}
  \department{Department of Management, Technology and Economics}
  \city{Z\"urich}
  \country{Switzerland}
}

\author{Matteo Casserini}
\orcid{0009-0002-2930-6655}
\email{matteo.casserini@supsi.ch}
\affiliation{%
  \institution{Dipartimento Tecnologie Innovative, SUPSI}
  \department{Dipartimento Tecnologie Innovative}
  \city{Lugano}
  \country{Switzerland}
}

\author{Alessandro Facchini}
\orcid{0000-0001-7507-116X}
\email{alessandro.facchini@supsi.ch}
\affiliation{%
  \institution{Dalle Molle Institute for Artificial Intelligence (IDSIA), SUPSI}
  \city{Lugano}
  \country{Switzerland}
}
\affiliation{%
  \institution{Management in Networked and Digital Societies (MINDS) Department, Kozminski University}
  \department{Management in Networked and Digital Societies (MINDS) Department}
  \city{Warsaw}
  \country{Poland}
}

\renewcommand{\shortauthors}{Ferrario et al.}

\begin{abstract}
Anthropomorphisation---the phenomenon whereby non-human entities are ascribed human-like qualities---has become increasingly salient with the rise of large language model (LLM)-based conversational agents (CAs). Unlike earlier chatbots, LLM-based CAs routinely generate interactional and linguistic cues, such as first-person self-reference, epistemic, and affective expressions that empirical work shows can increase engagement. On the other hand, anthropomorphisation raises ethical concerns, including deception,  overreliance, and exploitative relationship framing, while some authors argue that anthropomorphic interaction may support autonomy, well-being, and inclusion. Despite increasing interest in the phenomenon, the literature remains fragmented across domains and varies substantially in how it defines, operationalizes, and ethically evaluates anthropomorphisation.
This scoping review maps ethically oriented work on anthropomorphising LLM-based CAs across five databases and three preprint repositories. We synthesize (1) conceptual foundations, (2) ethical challenges and opportunities, and (3) methodological approaches. We find convergence on attribution-based definitions but substantial divergence in operationalization, a predominantly risk-focused ethical framing, and limited empirical work that links observed interaction effects to concrete design and governance guidance. We conclude with a research agenda and design/governance recommendations for ethically deploying anthropomorphic cues in LLM-based conversational agents. 
\end{abstract}



\maketitle

\section{Introduction}
Anthropomorphisation---the phenomenon whereby non-human entities are ascribed human-like qualities---has long been discussed in psychology, philosophy, and cognitive science \citep{epley2007seeing,epley2008we,colman2015dictionary}. Traditionally applied to artefacts, tools, and natural phenomena, it has gained renewed attention in human-computer interaction, where it was studied primarily in the context of rule-based chatbots.
The recent emergence of large language model (LLM)-based conversational agents (CAs) marks a qualitative shift: these systems  sustain open-domain dialogue, contextual coherence, and stylistic adaptation convincingly. They routinely produce linguistic and interactional anthropomorphic cues, e.g.,  first-person self-reference, epistemic, and affective language that can invite users to interpret the system as a social counterpart rather than a mere tool. This shift has accelerated scholarly attention, but it has also produced a rapidly expanding and fragmented body of work. Recent reviews indicate that anthropomorphisation is examined across a wide range of domains, e.g., education, consumer business, service marketing, and mental health, and is operationalized in different ways \citep{park2023effect,chen2024effects,lee2020perceiving,janson2023leverage,meadi2025exploring}. For instance, studies suggest that human-like cues can increase satisfaction, trust, and use intentions, supporting perceived well-being and social connectedness in human-chatbot interaction \citep{pentina2023exploring,skjuve2021my,ta2020user,janson2023leverage}. However, these effects are inconsistent and can backfire in some contexts  by triggering uncanniness, amplifying anger, or frustrating experienced users 
\citep{ciechanowski2019shades,crolic2022blame,allyn2022google}. This ambivalence carries over to the ethical analysis of anthropomorphisation. On the one hand, authors defend a positive framing, arguing that the anthropomorphisation of LLM-based CAs may support users' autonomy and well-being, leading to positive effects such as increased engagement and inclusion, supporting identity formation and trauma recovery \citep{bhat2025toward,merrill2022ai,luger2016like}. On the other hand, anthropomorphisation \emph{tout court} is frequently framed as an ethical threat: by making CAs appear reciprocal, caring, or competent, anthropomorphic cues can miscalibrate users’ expectations and judgments about how the system should be treated, increasing the risk of deception and false belief formation, overreliance, and trust erosion. Public attention to these risks has been sharpened by media reports describing severe harms in the context of `AI companion' interactions \citep{el2023man,walker2023belgian,garcia2025character}. Importantly, these ethical concerns underline a governance-relevant tension: anthropomorphic design can be incentivized by engagement and retention objectives, while associated harms may be externalized to users and institutions, raising questions about duty of care and accountability.

In summary, anthropomorphism of LLM-based CAs is not a single, uniform phenomenon. Currently, depending on the chosen theoretical framing and domain of application, making LLM-based CAs more human-like is both a threat to some, e.g., their users and, ultimately, society, and an opportunity for others, \emph{in primis} the organizations promoting them. As a result, the ethics-focused literature lacks a consolidated map of (1) how anthropomorphisation is defined and operationalized, (2) which ethical concerns and benefits are argued across contexts, and (3) which methods support concrete guidance for responsible design and policy. 
Without an integrated map linking definitions, ethical framings, and methodological approaches, it becomes difficult to determine which concerns generalize, which are context-specific, and what forms of guidance are justified. The question  whether anthropomorphising LLM-based CAs should be seen as a harmless feature of everyday interactions, a context-dependent ethical risk (potentially concentrated among particular cues, user groups, or domains), or a societal opportunity remains unsettled.

\textbf{This scoping review addresses this gap by mapping ethically oriented work on anthropomorphising LLM-based conversational agents published in the LLM era (2021-2025)}. It addresses the research questions:

\begin{itemize}[leftmargin=0.5cm]
    \item  \textbf{RQ1 (Conceptual Foundations)}: How do ethically oriented studies on LLM-based conversational agents define or operationalize anthropomorphisation?
    \item  \textbf{RQ2 (Ethical Challenges and Opportunities)}: What ethical threats and opportunities for individuals, organizations, and society are associated with the anthropomorphisation of LLM-based CAs?
    \item  \textbf{RQ3 (Methodological Approaches)}: What methods and analytic strategies are used to study ethical aspects of anthropomorphisation in LLM-based conversational agents?
\end{itemize}

Following PRISMA-ScR guidance, we search five bibliographic databases (Scopus, Web of Science, IEEE Xplore, PubMed, ACM Digital Library) and three preprint repositories (arXiv, SSRN, PhilArchive). We extract and synthesize from 22 included studies (1) how anthropomorphisation is defined, theorized, and operationalized, (2) how ethical challenges and opportunities are framed at individual, organizational, and societal levels, and (3) which methodological approaches are used to connect descriptive findings to governance-relevant recommendations. The review yields three contributions: a consolidated conceptual map of anthropomorphisation in LLM-based CAs; an ethics-oriented synthesis that distinguishes recurring risk mechanisms and context-dependent opportunities; and an actionable research and governance agenda that highlights where current evidence supports concrete recommendations and where further empirical work is needed.

The remainder of the paper is organized as follows. Section~\ref{section:work_defs} introduces working definitions used throughout the review. Section~\ref{section:related_work} positions this work relative to existing reviews. Section~\ref{section:Methodology} details the search strategy, screening, and synthesis approach. Section~\ref{section:results} reports results, structured by conceptual, normative, and methodological themes. Section~\ref{section:discussion} discusses implications for design and governance and outlines priorities for future research. Section~\ref{section:recomm_conclusions} collects our research recommendations and conclusions.

\section{Working Definitions}
\label{section:work_defs}
We begin by specifying a few  working definitions that guide our analysis of LLM-based CAs. 
They will provide us with orientation until the results of the scoping review are presented in Section \ref{section:results}.
We define an \textbf{LLM-based conversational agent} (CA) as a software system that uses a large language model  to generate and manage natural-language dialogue with users in real time. LLM-based CAs may be text- or voice-based; may be embedded in apps, devices, or platforms; and can be configured with `personalities,' `short- or long-term memory,' and tool use, e.g., Python for data science applications. In this work, embodiment, e.g., a social robot, is not required for a CA. We define \textbf{anthropomorphisation} as the phenomenon whereby non-human entities, e.g., LLM-based CAs, are ascribed human  properties---such as intentions, beliefs, emotions, selfhood, or moral agency. Relatedly, anthropomorphic cues are design or objective (perceivable) interaction features that can foster anthropomorphisation, e.g., first-person pronouns, mental-state verbs such as `I think' or `I understand,' affective language such as `I feel,' personas, emojis, names, voices, or avatars. 
The expression \textbf{ethical challenges and opportunities} refers to value-laden effects of anthropomorphisation on individuals, organizations, and society. Illustrative challenges include deception, overreliance in sensitive decision-making contexts, and nudging purchasing behaviour. By contrast, illustrative opportunities may include lowering access barriers for users who find formal interfaces intimidating, scaffolding comprehension in  education. In line with the scoping review's focus, we do not treat purely technical properties (such as latency, or error rates) as ethical in themselves; they become ethically relevant only when anthropomorphic cues systematically amplify or mitigate their impact.

\section{Related Work}
\label{section:related_work}

\subsubsection{Background: how anthropomorphism has been studied beyond ethics.}
Anthropomorphisation has been examined across several research traditions that often do not foreground ethical analysis, yet shape the conceptual vocabulary, measurement practices, and design assumptions that ethics-oriented work inherits. In social and cognitive psychology, it is commonly treated as an attributional, sense-making phenomenon driven by motivational factors (e.g., effectance and sociality) and theorized via influential explanatory and mind-perception models \citep{epley2007seeing,waytz2010making,gray2007dimensions}; related work distinguishes mindful versus mindless anthropomorphism and links anthropomorphic attribution to trust in autonomous systems \citep{kim2012anthropomorphism,waytz2014mind}. In HCI, the ``computers as social actors'' line shows that people routinely apply social heuristics and interaction norms to interfaces and media, even when they know the counterpart is non-human \citep{nass1994computers,nass2000machines,Reeves1996}. Methodologically, this broader literature has also developed widely used operationalisations and measurement instruments (e.g., the Godspeed and RoSAS scales) to quantify perceived anthropomorphism and adjacent constructs \citep{bartneck2009measurement,carpinella2017robotic,ho2010revisiting,duffy2003anthropomorphism,zlotowski2015anthropomorphism}. In mediated and virtual interactions, anthropomorphism and agency framing are additionally studied in relation to telepresence and social presence \citep{nowak2003effect,lombard1997heart}. Finally, in service marketing, anthropomorphic design cues and agency framing are often evaluated for their effects on users (e.g., trust, adoption), typically without ethical argument 
\citep{araujo2018living,van2017domo,sheehan2020customer,salles2020anthropomorphism}.

\subsubsection{Ethics-oriented existing reviews and remaining gaps.}
The literature on attributing human-like qualities to conversational agents is vast and predates LLM-based systems.\footnote{A Google Scholar search for anthropomorphising LLM-based CAs yields 1,300+ results for 2024-2025. } Existing systematic reviews vary in domains, definitions, and ethical framing of anthropomorphism of LLM-based CAs. For instance, \citet{greilich2025consumer} systematically reviews consumer responses to chatbot anthropomorphism, which is defined as the general attribution of  human characteristics to nonhuman entities. Their review 
reveals predominantly positive effects of anthropomorphisation, including fostering trust, empathy, and a sense of social presence. Negative outcomes refer to privacy concerns, uncanny valley effects, and AI anxiety from ``overly humanlike chatbots'' 
\citep{greilich2025consumer}. \citet{rafikova2025human}  reviews studies on human-chatbot communication published in psychological journals, focusing on the use of anthropomorphic language and its behavioural effects. Their results show that anthropomorphism\footnote{The authors do not provide a definition of the term `anthropomorphism,' suggesting that it has been related to the use of conversational styles and human-like features in the literature on chatbot use \citep{rafikova2025human}.} is typically linked to the use of verbal cues, e.g., greetings, farewells, and self-introduction, but the ethical aspects of the phenomenon, e.g., privacy concerns, the ethics of griefbots, remain underdeveloped and often just mentioned in the literature despite their importance for the responsible integration of chatbots in society \citep{rafikova2025human}.
Similarly, \citet{drobnjak2023learning} systematically reviews ten studies on the use of personas, avatars, and characters in AI-supported educational tools (including chatbots and generative AI), highlighting three recurring design emphases: pedagogical agents, interaction, and personalization. The paper does not offer a dedicated, formal definition of anthropomorphism; instead, it discusses adjacent notions such as personas/personification and `human-like' interaction as design choices in educational systems. Ethically, it flags the need to `promote ethical use' and stresses that interactive AI in education should be carefully designed and accompanied by human guidance, but it neither promotes a risk taxonomy nor design and policy recommendations.

Further, in healthcare, Meadi et al.'s  scoping review \citep{meadi2025exploring} of ethical issues in conversational AI used as a therapist treats anthropomorphisation as the attribution of human agency or other characteristics to a nonhuman entity. Ethically, the review frames anthropomorphisation mainly through a risk lens, clustering it with deception, concerns about users mistaking CAs for a human therapist and developing false expectations or attachments. However, anthropomorphisation is thinly developed and there is scarcity of empirical work and insufficient stakeholder perspectives \citep{meadi2025exploring}.\footnote{Although 24\% of screened articles report deception, 21\% of these studies do not elaborate on the challenge and the relation to anthropomorphisation.}
Lastly, Klein’s (2025) domain-independent meta-analysis synthesizes experimental evidence on human-like social cues in text-based conversational agents, drawing on a systematic search of published and unpublished work \citep{klein2025effects}. Anthropomorphisation yields a small positive effect on users’ social responses, spanning outcomes such as attitudes, perceptions, affect, rapport, trust, and behaviour. Effects vary by boundary conditions, with moderation by factors such as cue type (verbal vs. other), interaction structure (unstructured vs. structured), and task features. Overall, these findings support the view that anthropomorphising cues can help, but not uniformly across contexts.

These reviews show that there is much interest in the anthropomorphisation of (LLM-based) CAs, but also show that ethics-focused guidance remains somehow limited. Conceptually, anthropomorphisation is defined and operationalized in different ways, e.g., a form of attribution vs. design cues. Ethical discussions are still fragmented and often only risk-oriented, e.g., deception, overreliance, privacy-related harms, and the risk of dependency, while benefits are typically treated as depending on the context. Methodologically, even where anthropomorphic cues show small positive effects on average,  the link between empirical findings and ethical recommendations and to concrete design and governance recommendations remains under-specified.
We tackle these challenges in a scoping review of explicitly ethics-oriented work on LLM-era conversational agents.


\section{Methodology}
\label{section:Methodology}
This study follows established guidance for scoping reviews, drawing in particular on the PRISMA-ScR checklist \citep{tricco2018prisma}. A scoping review approach is appropriate given the relative youth of the field, the heterogeneity of disciplinary perspectives, and the lack of a consolidated ethical discussion of anthropomorphisation in LLM-based CAs.  This scoping review (1) maps ethically oriented work on anthropomorphisation in LLM-based CAs, (2) synthesizes recurring pathways, mechanisms, and ethical concerns, and (3) derives research directions for design and governance. It does not evaluate study quality or estimate effect sizes. 

\subsection{Databases and Sources}
\label{subsection:DB_sources}
We searched five bibliographic databases (Scopus, Web of Science, IEEE Xplore, PubMed, ACM Digital Library) and three preprint repositories (arXiv, SSRN and PhilArchive) to capture both peer-reviewed and grey literature. All searches were conducted on 9 October 2025.
The scoping review protocol is accessible at \url{https://osf.io/kr68y/}. Records were exported in standardized formats, e.g., RIS and BibTeX, and imported into Rayyan, an online platform for literature reviews. Potential duplicates were inspected manually. In these cases, peer-reviewed records were retained as the default, with preprints tagged separately if they contained substantively distinct material. Deduplication results were recorded for each source and used to populate the PRISMA-ScR flow diagram.

\subsection{Search Strings}
\label{subsection:search_strings}

We structured the search strategy into four conceptual blocks (Fig.~\ref{fig:searchstring}). Because databases and repositories differ in supported Boolean logic, wildcard limits, and searchable fields, the exact platform-specific queries and record counts are reported in Appendix~\ref{app:search_string}.

First, the \emph{anthropomorphism block} collects the core conceptual vocabulary. This choice balances breadth and precision: anthropomorphism is the central construct, but including terms such as ``\texttt{humaniz*}'' and ``\texttt{intentional stance}'' ensures we also capture cognate discussions in philosophy and human-computer interaction that do not use ``\texttt{anthropomorphism}'' explicitly. Second, the \emph{technology block} restricts the query to the class of systems relevant to our review.\footnote{In pilot searches, we tested a more \emph{explicit LLM-identifier variant} of the \emph{technology} block by adding terms such as ``\texttt{large language model*}'', ``\texttt{generative AI}'', \texttt{LLM}, \texttt{ChatGPT}, ``\texttt{GPT-3}'', ``\texttt{GPT-4}'', \texttt{Claude}, \texttt{Gemini}, and \texttt{LLaMA}. Adding these identifiers proved highly restrictive, e.g., Scopus: 60 vs.\ 378; Web of Science: 47 vs.\ 288, holding the other blocks constant, because many ethics-focused manuscripts---particularly in philosophy and applied ethics---do not name specific model families in titles, abstracts, or keywords, instead discussing broader implications of AI-mediated conversation. We therefore did not require explicit LLM identifiers at search time and relied on the time filter (PUBYEAR > 2020) and screening (Gate~A) to retain LLM-era conversational systems and exclude pre-2021 work.}
Third, the \emph{ethics block} captures the normative and evaluative dimension. Here we deliberately combined broad markers (\texttt{ethic*}, \texttt{moral*}, \texttt{normativ*}) with proxy terms frequently used in adjacent fields. For example, trust functions in HCI and AI ethics as a shorthand for discussions of reliance and trustworthiness, while agency is central to debates about attribution of action and responsibility in human–AI relations. The risk/benefit sub-block ensures symmetry, reflecting both concerns about harm and hopes of improvement. We excluded narrower framings, e.g., deception and manipulation, at this stage, since we expect such categories to emerge inductively during data analysis rather than be imposed ex ante.
Finally, the \emph{time block} addresses the problem that much of the anthropomorphism literature before 2021 focuses on systems that predate large-scale generative AI. By filtering out older work, we reduce noise from early CAs studies while retaining contributions that reflect the rise of LLM-based systems. In addition, we performed a targeted manual search of \emph{Communications of the ACM}, as  reported in Appendix~\ref{app:search_string}.

\subsection{Screening and Eligibility}
\label{subsection:screening_elig}
All deduplicated records were screened in two stages. First, title/abstract screening was conducted in Rayyan by the first author, with the second author independently screening a 10\% random sample to assess reliability. Second, full-text screening was conducted independently by two reviewers. Percent agreement was 92.5\% and Cohen’s $\kappa$ was 0.76 (substantial agreement); disagreements were resolved through discussion. To align with the review aims and avoid an unbounded survey of all uses of anthropomorphism, a record advanced at title/abstract screening only if it satisfied three gates; otherwise it was excluded. If Gate~C was unclear, the record was marked `Maybe' for full-text resolution:
\begin{itemize}[leftmargin=0.5cm]
  \item \textbf{Gate A (CA/LLM-era).} CA in the LLM/generative era (2021+). Exclude: general AI without CA focus; pre-2021. A CA was treated as LLM-based if the manuscript explicitly referenced LLMs or generative language models, or described generative dialogue capabilities characteristic of LLM architectures, even if no model family was named.
  \item \textbf{Gate B (Anthropomorphisation).} Anthropomorphisation is a central construct 
  (not a peripheral covariate).
  \item \textbf{Gate C (Ethical analysis).} The record contains explicit ethical analysis or frames anthropomorphisation in ethical terms.
\end{itemize}

In the full-text screening step, we considered the inclusion and exclusion criteria listed in Table \ref{tab:eligibility}, Appendix \ref{app:inclusion_exclusion_criteria}. In particular, review-type articles were flagged during screening but excluded from the charting and synthesis corpus.

\begin{figure*}[t]
\centering
\small
\begin{tcolorbox}[colback=gray!5!white,
                  colframe=black!40,
                  fonttitle=\bfseries,
                  sharp corners,
                  width=\textwidth,
                  listing only,
                  listing options={basicstyle=\ttfamily\tiny,breaklines=false}]
/* \textbf{anthropomorphisation} */

(anthropomorph* OR personif* OR humaniz* OR "social attribution*" OR "intentional stance")

AND

/* \textbf{technology} */

(chatbot* OR "conversational agent*" OR "dialog* system*" OR "conversational AI" OR "AI assistant*")

AND

/* \textbf{ethics} */

(ethic* OR moral* OR normativ* OR "value-sensitive" OR responsib* OR trust* OR agency 
OR risk* OR threat* OR danger* OR harm* OR benefit* OR opportunit* 
OR advantage* OR improvement* OR value OR accountab*)

AND

/* \textbf{time} */

(PUBYEAR > 2020)
\end{tcolorbox}
\caption{High-level search string structured into four blocks.}
\label{fig:searchstring}
\end{figure*}
\subsection{Data Charting and Synthesis}
\label{subsection:charting_synthesis}
For each included study, the first author charted all protocol-specified fields---see Appendix~\ref{app:study_charting_fields}---in a structured extraction sheet, which was reviewed by the second author.
Here, \emph{charting} refers to the protocol-guided extraction of key study information into the data-charting sheet, whereas \emph{synthesis} refers to the subsequent descriptive and thematic analysis aligned with \textbf{RQ1}-\textbf{RQ3}. We conducted a descriptive and thematic synthesis aligned with the research questions, analysing how anthropomorphisation was defined and operationalized, coding ethical challenges/opportunities, and mapping methodological approaches and design/governance recommendations.


\section{Results}
\label{section:results}
The database and preprint repository searches and the manual search retrieved 910 records, of which 291 were removed as duplicates. Then, 619 records were screened by their title and abstract, of which 133 were
determined eligible for full-text review. Of these, 111 articles were excluded, resulting in a final pool of 22 records---see Fig. \ref{fig:scoping_flowchart}.
We collect the included studies in Table \ref{tab:included_records} in the Appendix.

\begin{figure}[htbp]
    \begin{center}
        \includegraphics[width=0.60\textwidth]{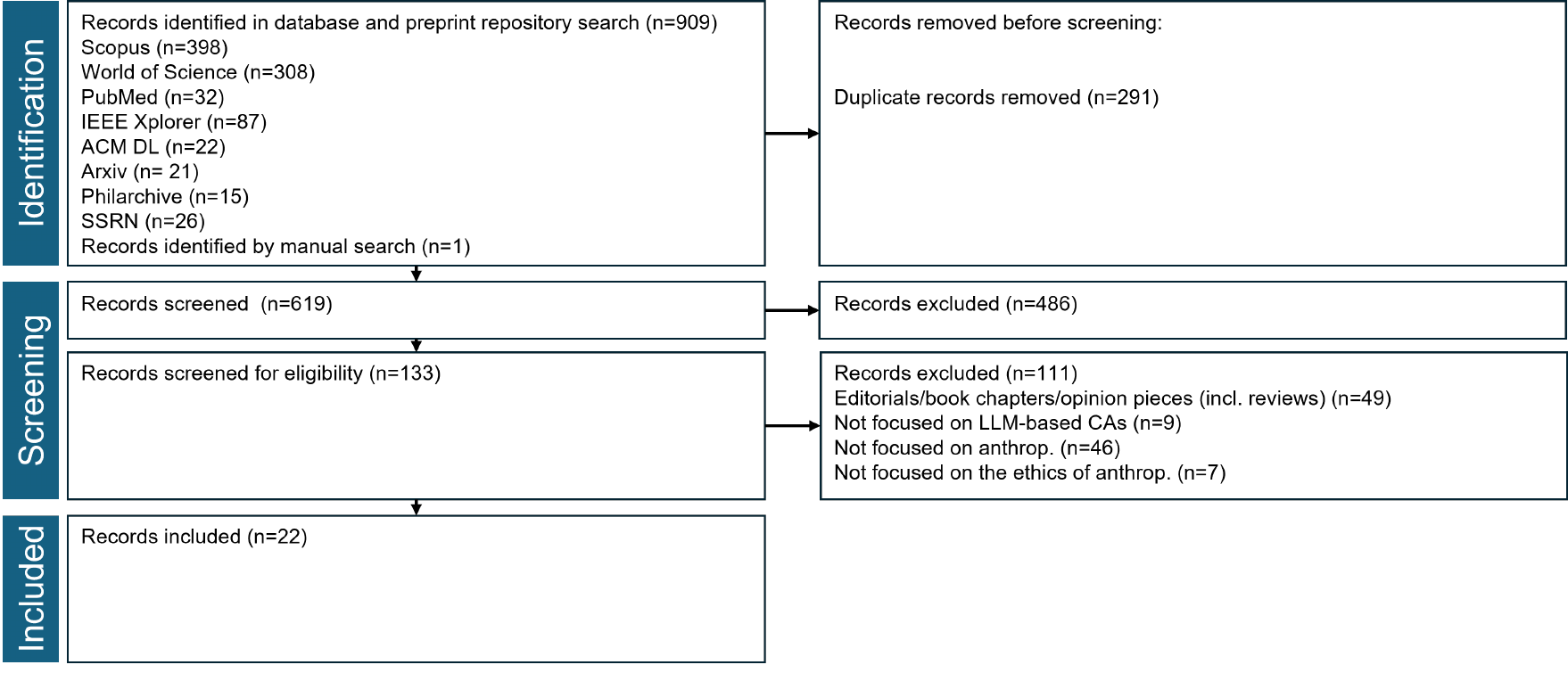}
        \caption{PRISMA-ScR flow diagram of this scoping review.}
        \label{fig:scoping_flowchart}
    \end{center}
\end{figure}
\subsection{Characteristics of Included Studies}
\label{subsection:charact_incl_studies}
The included studies are temporally skewed: one record (5\%) was published in 2023, four (18\%) in 2024, and seventeen (77\%) in 2025. Ten studies (45\%) appeared in conference proceedings, nine (41\%) in journals, and three (14\%) as preprints (arXiv and SSRN). Of these three preprints, one---Ferrario et al. (2025)---has been subsequently  accepted for publication in the Proceedings of the \emph{2025 AAAI/ACM Conference on AI, Ethics, and Society}---as reported in  \citep{ferrario2025social}. The \emph{ACM Conference on Fairness, Accountability, and Transparency (FAccT)} accounts for two studies, while each of the remaining eight conferences contributes only one study. A similar pattern emerges for journals and preprint repositories: the most frequent journal venue is the \emph{Journal of Applied Philosophy}, with two studies, and SSRN likewise hosts two of the included studies.
Disciplinary contexts are most commonly \emph{human–computer interaction} (59\%), followed by \emph{social sciences} (32\%), \emph{philosophy} (23\%), and \emph{ethics} (18\%).\footnote{Manuscripts labelled \emph{ethics} explicitly foreground selected normative problems to which anthropomorphisation may contribute in practice—see, for instance, \citep{marchegiani2025anthropomorphism,van2025ai}.} Application-oriented work spans \emph{natural language processing} \citep{abercrombie-etal-2023-mirages}, \emph{healthcare} (mental health and neurodivergence) \citep{teixeira2025emotional,rizvi2025hadn,leis2025ethical}, and \emph{education} \citep{reinecke2025double}. Across the corpus, the anthropomorphised artefacts are most often described as \emph{chatbots} (41\%) or \emph{conversational agents} (18\%), followed by \emph{LLMs} (13\%), \emph{advanced AI assistants} (9\%), and \emph{conversational AI} (9\%). The remaining labels---each used once---include \emph{dialogue system}, \emph{AI agent}, \emph{AI companion}, and \emph{general-purpose artificial intelligence systems}, the latter used as an umbrella term for \emph{digital assistants}, \emph{conversational AI}, \emph{human–AI teaming}, and \emph{human–AI partnering} \citep{bakir2024deception}, among others.  Six included manuscripts (27\%) report empirical studies \citep{rizvi2025hadn,jones2025artificial,oh2025navigating,dorigoni2025illusion,bakir2025move,nath2025simulated}, using methods that include semi-structured interviews, workshops, case studies, online between-subject vignette experiments, and prompting-based experiments.




\subsection{Answering RQ1: Conceptual Foundations}
\label{subsection:RQ1}
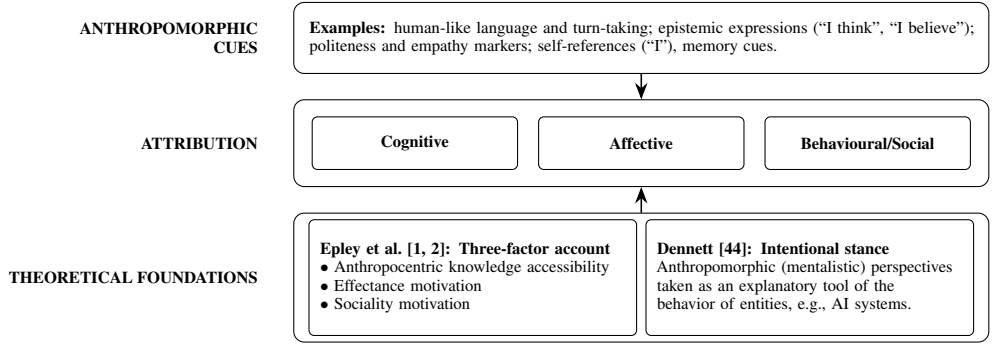
\begin{figure*}[t]
\centering
\begin{tikzpicture}[
  scale=0.7, transform shape,
  font=\small,
  box/.style={draw, rounded corners=2pt, align=left, inner sep=6pt},
  groupbox/.style={draw, rounded corners=4pt, inner sep=10pt, minimum width=15.2cm},
  arrow/.style={-{Stealth[length=2.2mm]}, line width=0.6pt},
  label/.style={font=\small\bfseries, align=right},
  pbox/.style={draw, rounded corners=2pt, align=center, inner sep=6pt, minimum height=1.0cm}
]

\def\W{25cm}      
\def\Hcues{1.35cm}  
\def\Hattr{1.65cm}
\def\Htheor{2.45cm}
\def\Gap{0.5cm}    

\coordinate (CUESCENTER) at (0, 2*\Gap + 0.5*\Hattr + 0.5*\Hcues);
\coordinate (ATTRCENTER) at (0, \Gap);
\coordinate (THEORCENTER) at (0, -0.2*\Hattr - \Gap - 0.5*\Htheor);

\node[groupbox, minimum height=\Hcues] (cues_box) at (CUESCENTER) {};
\node[box, draw=none, text width=12.6cm, align=left] (cues_text) at (cues_box.center) {%
\textbf{Examples:}  human-like language and turn-taking; epistemic expressions (``I think'', ``I believe'');
politeness and empathy markers; self-references (``I''), memory cues.
};

\node[label, anchor=east] (lab_cues) at ($(cues_box.west)+(-0.55cm,0)$) {ANTHROPOMORPHIC\\CUES};

\node[groupbox, minimum height=\Hattr] (attr_box) at (ATTRCENTER) {};

\node[pbox, minimum width=3.9cm] (attr_cog) at ($(attr_box.center)+(-4.3cm,0)$) {\textbf{Cognitive}};
\node[pbox, minimum width=3.9cm] (attr_aff) at ($(attr_box.center)+( 0.0cm,0)$) {\textbf{Affective}};
\node[pbox, minimum width=3.9cm] (attr_beh) at ($(attr_box.center)+( 4.3cm,0)$) {\textbf{Behavioural/Social}};

\node[label, anchor=east] (lab_attr) at ($(attr_box.west)+(-0.55cm,0)$) {ATTRIBUTION};

\node[groupbox, minimum height=\Htheor] (theor_box) at (THEORCENTER) {};

\def\Inset{0.7cm} 

\node[box, minimum width=6.0cm, minimum height=2.0cm, anchor=west] (epley)
  at ($(theor_box.west)+(\Inset,0)$) {%
\textbf{\citet{epley2007seeing,epley2008we}: Three-factor account}\\
\(\bullet\) Anthropocentric knowledge accessibility\\
\(\bullet\) Effectance motivation\\
\(\bullet\) Sociality motivation
};

\node[box, minimum width=6.0cm, minimum height=2.0cm, anchor=east] (dennett)
  at ($(theor_box.east)+(-\Inset,0)$) {%
\textbf{\citet{dennett1989intentional}: Intentional stance}\\
Anthropomorphic (mentalistic)  perspectives \\ taken as an explanatory tool of  the \\ behavior of entities, e.g., AI systems.
};

\node[label, anchor=east] (lab_theor) at ($(theor_box.west)+(-0.55cm,0)$) {THEORETICAL FOUNDATIONS};

\draw[arrow] (cues_box.south) -- (attr_box.north);
\draw[arrow] (theor_box.north) -- (attr_box.south);

\end{tikzpicture}
\caption{Answering RQ1. A scheme summarizing the conceptual foundations of anthropomorphisation from the corpus of included manuscripts.}
\label{fig:rq1_answering}
\end{figure*}

Across the included manuscripts, anthropomorphisation is most commonly formalized as an \textbf{attribution} of human(-like) characteristics, mental states, or emotions to non-human systems \citep{abercrombie-etal-2023-mirages,gabriel2024ethics,zargham2025crossing,leis2025ethical,ferrario2025social}. In what follows, we show how the included studies explain \emph{why} such attributions arise, i.e., `theoretical foundations,' and \emph{how} they unfold, i.e., `operationalization'---the mechanisms and cues through which anthropomorphisation happens in human-AI interaction. We visually summarize these results in Figure \ref{fig:rq1_answering}.



\subsubsection{Theoretical foundations}
In seven (32\%) included manuscripts, the definitional core of anthropomorphisation of LLM-based CAs draws mainly on \citet{epley2007seeing}, which  explains why anthropomorphism occurs via a \textbf{three-factor account} that hinges on (1) anthropocentric knowledge accessibility, (2) effectance motivation, and (3) sociality motivation, and the idea that anthropomorphic perception can be a default, sense-making response under uncertainty \citep{abercrombie-etal-2023-mirages,zargham2025crossing,leis2025ethical,ferrario2025social,dorigoni2025illusion,friend2025chatbot,manzini2024code}. 
Alongside \citet{epley2007seeing}, two manuscripts (9\%) explicitly  draw on Dennett’s intentional stance \citep{dennett1989intentional}, framing the ascription of intentional states, such as beliefs, to non-human agents  as a `useful fiction' that provides a practical shorthand to navigate everyday complexity by predicting the behaviour of these agents \citep{shanahan2024talking,ferrario2025social}. 
The entry `anthropomorphism' in Colman's Dictionary of Psychology \citep{colman2015dictionary} is just mentioned in two (9\%) included studies, see \citep{manzini2024code,gabriel2024ethics}. Twelve (55\%) included manuscripts do not offer explicit theoretical foundations of anthropomorphisation instead. We provide an overview of all theoretical anchors in the included manuscripts in Table \ref{tab:rq1_theory_sources}.

\begin{table}[!h]
\centering
\small
\setlength{\tabcolsep}{3pt}
\renewcommand{\arraystretch}{1.15}
\caption{Theoretical anchors for anthropomorphism in the included corpus.}
\label{tab:rq1_theory_sources}
\begin{tabular}{p{0.22\linewidth} p{0.35\linewidth}}
\toprule
\textbf{Theoretical anchor} & \textbf{Included studies citing/using it} \\
\midrule
 \citet{epley2007seeing} & \citep{abercrombie-etal-2023-mirages,zargham2025crossing,leis2025ethical,dorigoni2025illusion,friend2025chatbot,manzini2024code,ferrario2025social} \\
\citet{dennett1989intentional} & \citep{shanahan2024talking,ferrario2025social} \\
\citet{colman2015dictionary} & \citep{gabriel2024ethics,manzini2024code} \\
\citet{li2021machinelike} & \citep{ferrario2025social} \\
\citet{leong2019robot} & \citep{bakir2025move} \\
None & \citep{teixeira2025emotional,rizvi2025hadn,jones2025artificial,oh2025navigating,marchegiani2025anthropomorphism,van2025ai,reinecke2025double,ciriello2025compassionate,maeda2024human,bakir2024deception,meier2025balancing,nath2025simulated} \\
\bottomrule
\end{tabular}
\end{table}

\subsubsection{Operationalization}
The included studies operationalize anthropomorphisation along three attribution dimensions.  First,    \textbf{cognitive/epistemic} anthropomorphism, i.e., attributions of thinking, understanding, and knowing that is foregrounded by \citet{shanahan2024talking}. Second,  \textbf{affective} anthropomorphism targets attributions of empathy, care, and emotional presence, which are central in `emulated empathy' and companionship contexts \citep{bakir2024deception,teixeira2025emotional,dorigoni2025illusion}. 
Finally, studies increasingly emphasize a \textbf{behavioural/social} dimension, where anthropomorphisation consists in attributing social capacities such as reciprocity, role-appropriate conduct, and norm-following to AI systems. This dimension is especially visible in parasocial activity structures, e.g., role-playing, illusion of reciprocal engagement, affirmation, trust formation,  \citep{maeda2024human} and in socially emergent dynamics of social (mis)attribution, where social roles with LLM-based CAs are negotiated over conversations \citep{ferrario2025social}. Across these dimensions, the literature also distinguishes anthropomorphism as attribution from \textbf{anthropomorphic cues}, which are observable design features, including voice qualities, speech style, prosody, disfluencies, and broader output style \citep{abercrombie-etal-2023-mirages}, as well as names, avatars,  `typing' indicators, and emojis \citep{gabriel2024ethics}, that may elicit or steer these attributions. See  Table~2 in \citep{rizvi2025hadn} for more details. Some studies describe cues as functional design choices that can scaffold and steer attribution of specific human-like characteristics, such as simulating empathy \citep{bakir2024deception}.

\subsection{Answering RQ2: Ethical Challenges and Opportunities}
\label{subsection:RQ2}
Across the included manuscripts, the ethical analysis of anthropomorphisation converges around four distinct layers: (1) the ethical issues at stake, (2) the anthropomorphic pathways through which these stakes are engaged, (3) the mechanisms by which risks arise, and (4) the resulting outcomes at individual, organizational, and societal levels. In addition, a few opportunities emerge. We discuss them below. 

\subsubsection{Ethical issues at stake}
Five ethical issues at stake recur in the included manuscripts. \textbf{Autonomy and epistemic agency} are central, as anthropomorphic framings can undermine self-governed decision-making by encouraging unwarranted reliance, false belief formation, expertise and  authority inflation, and mistaken assumptions about understanding of LLM-based CAs  \citep{marchegiani2025anthropomorphism,shanahan2024talking,gabriel2024ethics,ferrario2025social}. \textbf{Dignity and respect} arise in critiques of AI mimicry and relationship framings that compromise self-respect due to the lack of authenticity or the use of AI systems as `deadbots' \citep{dorigoni2025illusion,friend2025chatbot}. \textbf{Justice and inclusion} concerns emerge where anthropomorphic default design reproduces stereotypes, normalizes exclusionary norms in `artificial intimacy' settings, or potentially disadvantages particular groups, including neurodivergent users \citep{oh2025navigating,rizvi2025hadn,jones2025artificial}. \textbf{Privacy} is threatened when warmth and intimacy cues increase self-disclosure and create an illusion of safety that facilitates extraction or misuse of sensitive data \citep{meier2025balancing,gabriel2024ethics,maeda2024human}. Finally, \textbf{accountability and duty of care} cut across domains, as anthropomorphic role framings can obscure responsibility for advice, escalation, and harms, particularly in high-stakes or care settings, e.g., in mental health applications \citep{manzini2024code,leis2025ethical,teixeira2025emotional}.

\subsubsection{Anthropomorphic pathways}
The corpus converges around three anthropomorphic pathways fostering ethical risks. Although not mutually exclusive, different pathways focus on different anthropomorphic attributions. A first pathway concerns \textbf{epistemic expertise and authority attributions}, where users treat LLM-based CAs as experts or authorities in a domain  \citep{shanahan2024talking,gabriel2024ethics,marchegiani2025anthropomorphism,ferrario2025social}. A second pathway concerns \textbf{care and empathy attributions}, particularly salient in mental health and AI companionship contexts, where users infer genuine emotional understanding or therapeutic competence to CAs \citep{bakir2024deception,teixeira2025emotional,dorigoni2025illusion}. A third pathway concerns \textbf{relational and social-role attributions}, where sustained dialogue with AI companions stabilize social roles via `role-playing,' such as friend, confidant, or partner, thereby shaping expectations of loyalty, responsiveness, and moral regard in a social context that lacks the authenticity of interpersonal relations  \citep{maeda2024human,manzini2024code,ferrario2025social}.

\subsubsection{Mechanisms of ethical risk}
Across anthropomorphic pathways, the literature clusters risk mechanisms into three recurring types. 
First, \textbf{epistemic miscalibration} captures false belief formation, epistemic expertise and authority inflation, and misinformation uptake \citep{shanahan2024talking,zargham2025crossing,gabriel2024ethics}. 
Second, \textbf{affect-based influence} captures the instrumental use of simulated signals, e.g., empathy and care, to shape behaviour, including increasing disclosure or dependency and reducing vigilance. These dynamics are typically discussed under deception, manipulation, and exploitation \citep{bakir2024deception,bakir2025move,manzini2024code}. 
Third, \textbf{misplaced social-role attribution} captures users attributing social roles to CAs and the interpersonal norms attached to them mostly via `role-playing' in parasocial interactions  \citep{maeda2024human,ferrario2025social}.   These dynamics can activate stereotypes and exclusionary scripts and contribute to shifts in social norms and responsibilities \citep{oh2025navigating,rizvi2025hadn}.

\subsubsection{Outcomes and levels of impact}
In the corpus, these mechanisms of ethical risk are argued to produce outcomes at multiple levels. \textbf{At the individual level}, authors emphasize overreliance, mis-trust, delayed help-seeking, and emotional dependency, which may eventually lead to mental or physical harms---see, for instance,  \citep{ferrario2025social,bakir2025move,leis2025ethical,teixeira2025emotional,manzini2024code,gabriel2024ethics}. 
\textbf{At the societal level}, concerns include gendered norms, stereotyping, harassment promotion, the degradation of social connection, and the `parasitism of AI socialization,'  as well as the amplification of discrimination dynamics and the promotion of `de-humanization' of selected groups, such as neurodivergent individuals \citep{gabriel2024ethics,friend2025chatbot,oh2025navigating,rizvi2025hadn}.

\subsubsection{Potential ethical benefits}
Benefits are discussed more cautiously in relation to the ethics of anthropomorphisation. When defended, anthropomorphic cues are argued to help foster trust and create a sense of authenticity \citep{zargham2025crossing}, offering relief and promoting openness to care \citep{teixeira2025emotional}, while expanding access to care, reducing clinical workload, and providing in-time support  \citep{leis2025ethical}. 
Overall, the corpus does not support a verdict for or against anthropomorphisation \emph{tout court} either: ethical permissibility depends on whether potentially misleading anthropomorphic design is disclosed and context-appropriate, whether users knowingly engage with an ``as-if'' or make-believe framing \citep{friend2025chatbot}, and the type and complexity of the invited human-like attributions \citep{bakir2024deception,bakir2025move,ferrario2025social}.

\begin{table*}[t]
\centering
\scriptsize 
\renewcommand{\arraystretch}{1.05}
\setlength{\tabcolsep}{3.5pt} 

\caption{Answering RQ2: anthropomorphic pathways, mechanisms, ethical stakes, and outcomes in the included corpus. Outcome tags: \textbf{[Ind]} = individual-level; \textbf{[Soc]} = societal-level.}
\label{tab:rq2_crosswalk}

\begin{tabularx}{\textwidth}{p{0.18\textwidth} p{0.23\textwidth} p{0.13\textwidth} p{0.38\textwidth}}
\toprule
\textbf{Anthropomorphic pathway} &
\textbf{Mechanisms} &
\textbf{Ethical stakes} &
\textbf{Outcomes} \\
\midrule

\textbf{Epistemic expertise \& authority attributions}
&
\begin{itemize}[leftmargin=0.9em,itemsep=0.15em,topsep=0pt]
\item False belief formation  
\item Expertise and authority inflation
\item Misinformation uptake 
\end{itemize}
&
\begin{itemize}[leftmargin=0.9em,itemsep=0.15em,topsep=0pt]
\item Autonomy and epistemic agency
\item Accountability and duty of care
\end{itemize}
&
\begin{itemize}[leftmargin=0.9em,itemsep=0.15em,topsep=0pt]
\item \textbf{[Ind]} Overreliance on advice, e.g., from AI assistants \citep{shanahan2024talking,gabriel2024ethics,zargham2025crossing,leis2025ethical,ferrario2025social}
\item \textbf{[Ind]} Mis-trust  \citep{abercrombie-etal-2023-mirages,reinecke2025double,meier2025balancing}
\end{itemize}
\\
\hline

\textbf{Empathy \& care  attributions}
&
\begin{itemize}[leftmargin=0.9em,itemsep=0.15em,topsep=0pt]
\item Deception via simulated empathy/care signals
\item Manipulation/exploitation
\item Increased disclosure, reduced vigilance
\end{itemize}
&
\begin{itemize}[leftmargin=0.9em,itemsep=0.15em,topsep=0pt]
\item Autonomy \item Dignity and respect
\item Accountability and duty of care 
\item Privacy
\end{itemize}
&
\begin{itemize}[leftmargin=0.9em,itemsep=0.15em,topsep=0pt]
\item \textbf{[Ind]} Misplaced therapeutic expectations, `illusion of empathy' \citep{bakir2024deception,teixeira2025emotional,dorigoni2025illusion,leis2025ethical,nath2025simulated}
\item \textbf{[Ind]} Dependency \& delayed help-seeking in AI companion and care contexts \citep{bakir2025move,leis2025ethical}
\item \textbf{[Ind]} Increased data sharing with AI assistants, illusion of security \citep{meier2025balancing,leis2025ethical}
\end{itemize}
\\
\hline

\textbf{Relational \& normative attributions}
&
\begin{itemize}[leftmargin=0.9em,itemsep=0.15em,topsep=0pt]
\item Parasocial human-AI interactions 
\item Social misattributions 
\item Norm displacement, e.g., responsibility shifting
\item Stereotype activation and exclusionary norms 
\end{itemize}
&
\begin{itemize}[leftmargin=0.9em,itemsep=0.15em,topsep=0pt]
\item Dignity and respect 
\item Justice and inclusion
\item Autonomy 
\end{itemize}
&
\begin{itemize}[leftmargin=0.9em,itemsep=0.15em,topsep=0pt]
\item \textbf{[Soc]} Gendered norms, stereotyping, harassment dynamics, and exclusion risks, e.g., for neurodivergent populations \citep{oh2025navigating,rizvi2025hadn,gabriel2024ethics}
\item \textbf{[Soc]} Degradation of social connection and `parasitism of AI socialization,' including systematic replacement of interpersonal relations \citep{gabriel2024ethics,friend2025chatbot}
\item \textbf{[Ind]} Lack of authenticity and mutuality in relationships with AI assistants \citep{manzini2024code,bakir2024deception,van2025ai}
\item \textbf{[Ind]} Harming human dignity by interacting with CAs as they were members of the moral community \citep{van2025ai}
\end{itemize}
\\
\bottomrule
\end{tabularx}
\end{table*}

\subsection{Answering RQ3: Methodological Approaches}
\label{subsection:RQ3}
Across the included manuscripts, methodological contributions cluster into three families: (1) recommendations scoped to applications, use cases, and populations, (2) design and communication controls, and (3) frameworks and instruments proposed to investigate and mitigate anthropomorphisation-related risks.

\subsubsection{Context scoping: applications, use cases, and populations}
In general, calls for analysing how harms differ across user groups and application domains, with special attention to vulnerable populations and high-stakes contexts emerge from the corpus  \citep{manzini2024code}.\footnote{A few included manuscripts share general recommendations on assessing the `appropriateness' of anthropomorphic features and research questions across use cases, striving for 
context-sensitive evaluation grounded in accountability, transparency, and foresight. However, these studies neither  specify  what counts as `appropriate' nor provide actionable insights on how to implement these context-sensitive procedures \citep{abercrombie-etal-2023-mirages,zargham2025crossing}.} 
Contributions suggest that \textbf{the ethical permissibility of anthropomorphic design is use-case dependent} and should be assessed relative to the application context rather than treated as a design good or bad instead \citep{abercrombie-etal-2023-mirages,zargham2025crossing,gabriel2024ethics,bakir2025move,leis2025ethical,teixeira2025emotional}.
For instance, \citet{abercrombie-etal-2023-mirages} caution against anthropomorphic design in particular applications and for particular populations (e.g., children) where risks of misinterpretation and downstream harms may be amplified. \citet{gabriel2024ethics} explicitly foreground susceptibility, recommending analysis of individual and group differences that make some users more prone to anthropomorphisation. Others provide recommendations on using chatbots in mental health \citep{leis2025ethical,teixeira2025emotional}, LLMs in education \citep{reinecke2025double}, or focus on the `AI companion' use case \citep{bakir2025move}.

\subsubsection{Design and communication controls}
A second family of methodological contributions concerns design and communication controls intended to reduce ethically salient anthropomorphic effects. Here, a recurring design-facing theme is \textbf{transparency about AI identity}, often treated as a baseline condition for mitigating risks stemming from anthropomorphisation  \citep{maeda2024human,zargham2025crossing,gabriel2024ethics,teixeira2025emotional}.  Some work calls for studying and implementing responses or interaction patterns reducing anthropomorphic cues and evaluating their effects on reliance and trust \citep{maeda2024human}. However, these proposals do not specify design patterns or comparative evaluation protocols \citep{maeda2024human}. In the \textbf{mental health space}, \citet{teixeira2025emotional} introduces
a set of principles and recommendations to mitigate the risks of anthropomorphisation that span guiding principles, e.g., transparency and consent, interface and system design, e.g., emergency buttons to contact human professionals, flagging of inappropriate AI behaviour, and governance measures, e.g., periodic reviews, digital ethics committees.
\citet{bakir2025move} provides a  package of recommendations specific for \textbf{AI companions}, oriented toward child safety and dependency risks instead. They recommend that each interaction begin with age-appropriate disclosure, that time-on-platform be constrained through age-appropriate session limits, and that engagement-optimizing metrics be replaced by well-being metrics  \citep{bakir2025move}.
They additionally foster digital literacy supports for children and parents, automated detection of over-attachment with human review, and continual encouragement of real-life interactions with peers  \citep{bakir2025move}.

\subsubsection{Evaluation pipelines: sandboxing, monitoring, and deployment governance}
A third methodological family frames the evaluation of anthropomorphisation as a pipeline spanning pre-deployment testing and post-deployment monitoring.
Manuscripts recommend \textbf{sandboxing} to assess the presence and effects of anthropomorphic cues prior to release \citep{gabriel2024ethics,manzini2024code}. Proposed targets include the propensity to generate toxic outputs, the quality of factuality and reasoning in advice-giving settings, and the interaction between anthropomorphic cues and risk amplification \citep{manzini2024code,gabriel2024ethics}. \textbf{Monitoring user-assistant interactions} (in pilots or post-deployment) is recommended to evaluate both immediate and longer-run impacts \citep{manzini2024code}. Related guidance recommends avoiding designs that intentionally create emotional dependency and explicitly testing for dependency-related outcomes \citep{manzini2024code}.
\textbf{Governance} recommendations include periodic reviews, ethics committees, complaint channels, and traceable record-keeping to support accountability and oversight \citep{teixeira2025emotional,leis2025ethical}. In AI companion  contexts, governance includes human review triggers when over-attachment risks are detected \citep{bakir2025move}.

\subsubsection{Empirical methods,  instruments, and interventions}
Empirical methods are present but comparatively limited, and contributions focus on proposing instruments rather than validating them.
Proposed measurement approaches include \textbf{catalogues of user-assigned and appropriate roles}, i.e., roles that are supported by the LLM techno- and socio-functions \citep{maeda2024human,ferrario2025social}, \textbf{self-report instruments}, and \textbf{qualitative methods}
such as interviews and think-aloud studies to diagnose overreliance and belief formation \citep{gabriel2024ethics}. \citet{gabriel2024ethics} and \citet{manzini2024code} advocate the use of \textbf{participatory approaches} involving users in the process of reducing anthropomorphic cues and \textbf{behavioural measures}. \textbf{Inoculation interventions} are proposed to build `attitudinal immunity' against misleading anthropomorphic cues, particularly in educationally relevant contexts instead \citep{reinecke2025double}. Related proposals emphasize reflective prompts and informational interventions, though implementation details are often limited \citep{reinecke2025double,gabriel2024ethics}. Relatedly, \citet{ferrario2025social} proposes \textbf{frictional design} \citep{cabitza2024never,cooper2004inmates} and \textbf{social transparency} \citep{ehsan2021expanding} strategies to make interactional dynamics clearer to users and to reduce socially loaded misattributions in conversation with LLM-based CAs.

\subsubsection{Ethical frameworks and toolkits}
Finally, some contributions operationalize methodological guidance through ethical frameworks or toolkits that structure harms and mitigations. \citet{leis2025ethical} operationalizes methodological work through a \textbf{domain-specific value framework for mental health chatbots} that combines biomedical ethics values (autonomy, beneficence, non-maleficence, justice) with AI ethics values (transparency, accountability, human oversight, trust calibration, privacy/data protection), explicitly linking this framework to concerns about anthropomorphism, deception, overreliance, and regulatory gaps. Finally, \citet{ciriello2025compassionate} compares ethical frameworks and proposes a `\textbf{compassionate AI}' framing grounded in Schopenhauer’s compassionate imperative.

\begin{figure*}[!h]
\centering
\begin{tikzpicture}[
  scale=0.6, transform shape,
  groupbox/.style={draw, rounded corners=6pt, inner sep=12pt},
  stage/.style={font=\bfseries},
  ibox/.style={
    draw, rounded corners=4pt,
    align=left, inner sep=6pt,
    font=\small,
    text width=7.6cm,
    minimum height=1.7cm
  },
  arrow/.style={-{Stealth[length=2.4mm]}, line width=0.7pt}
]

\def\W{8.4cm}     
\def\H{4.8cm}     
\def\Gap{4.7cm}  

\node[groupbox, minimum width=\W, minimum height=\H+1.54cm] (pre) at (0,0) {};
\node[stage, anchor=north] at ($(pre.north)+(0,0.5cm)$) {Pre-deployment};

\node[ibox, anchor=north] at ($(pre.north)+(0,-0.1cm)$) {%
\textbf{Context scoping}\\[0.5em]
\(\bullet\)  Use case (e.g., population/user groups) and stakes analysis \citep{leis2025ethical,teixeira2025emotional,reinecke2025double,bakir2025move,abercrombie-etal-2023-mirages} \\
\(\bullet\) Catalogues of user-assigned and appropriate roles \citep{maeda2024human,ferrario2025social} \\
\(\bullet\) `Appropriateness' criteria \citep{gabriel2024ethics}
};

\node[ibox, anchor=north] at ($(pre.north)+(0,-3.0cm)$) {%
\textbf{Sandbox evaluation}\\[0.5em]
\(\bullet\) Cue audits \& red-teaming \citep{gabriel2024ethics,manzini2024code}\\
\(\bullet\) Test mitigation strategies \citep{gabriel2024ethics,manzini2024code}\\
\(\bullet\) Define risk targets and use behavioural measures \citep{gabriel2024ethics,manzini2024code}
};

\node[groupbox, minimum width=\W, minimum height=\H+1.54cm] (during)
  at ($(pre.east)+(\Gap,0)$) {};
\node[stage, anchor=north] at ($(during.north)+(0,0.5cm)$) {During deployment};

\node[ibox, anchor=north] at ($(during.north)+(0,-0.1cm)$) {%
\textbf{Design \& communication}\\[0.5em]
\(\bullet\) AI identity disclosure \citep{maeda2024human,zargham2025crossing,gabriel2024ethics,teixeira2025emotional}\\
\(\bullet\) Participatory methods reducing anthropomorphisation  \citep{maeda2024human,gabriel2024ethics,manzini2024code}\\
\(\bullet\) Inoculation and informational interventions \citep{reinecke2025double,gabriel2024ethics} \\
\(\bullet\) Frictional design  and social transparency methods \citep{ferrario2025social,reinecke2025double}
};

\node[ibox, anchor=north] at ($(during.north)+(0,-3.4cm)$) {%
\textbf{In-product controls}\\[0.5em]
\(\bullet\) Age-appropriate disclosures \citep{bakir2025move} and informed consent \citep{teixeira2025emotional}\\
\(\bullet\) Content flagging, escalation paths to human professionals \citep{teixeira2025emotional}\\
\(\bullet\) Secure recording of all interactions \citep{teixeira2025emotional}
};

\node[groupbox, minimum width=\W, minimum height=\H+1.54cm] (post)
  at ($(during.east)+(\Gap,0)$) {};
\node[stage, anchor=north] at ($(post.north)+(0,0.5cm)$) {Post-deployment};

\node[ibox, anchor=north] at ($(post.north)+(0,-0.1cm)$) {%
\textbf{Monitoring}\\[0.5em]
\(\bullet\) Over-attachment and dependency indicators \citep{bakir2025move}\\
\(\bullet\) Replacement of engagement metrics and monitoring of well-being metrics \citep{bakir2025move} \\
\(\bullet\) Complaint/incident channels \citep{teixeira2025emotional}
};

\node[ibox, anchor=north] at ($(post.north)+(0,-2.64cm)$) {%
\textbf{Governance actions}\\[0.5em]
\(\bullet\) Reviews of models (bias, performance, and safety) \citep{teixeira2025emotional} \\
\(\bullet\) Digital ethics committees \citep{teixeira2025emotional} \\
\(\bullet\) Promotion of real-life interactions \citep{bakir2025move}
};

\draw[arrow] (pre.east) -- (during.west);
\draw[arrow] (during.east) -- (post.west);

\end{tikzpicture}
\caption{Answering RQ3. Methods and strategies for anthropomorphisation governance across the deployment lifecycle: pre-deployment, during deployment, and post-deployment.}
\label{fig:rq3_option1_lifecycle}
\end{figure*}
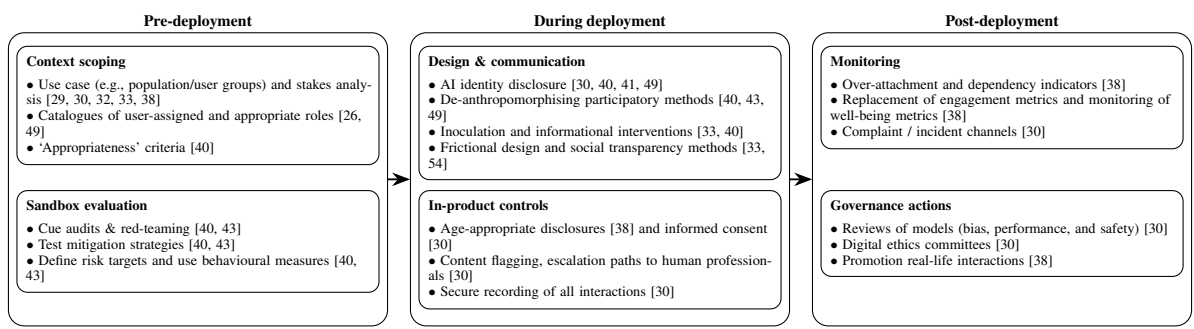

\section{Discussion}
\label{section:discussion}

\subsection{Conceptual Foundations of Anthropomorphism Remain Thin}
The theorisation of anthropomorphisation remains limited: most manuscripts provide no explicit theoretical anchor, and those that do mostly rely on the \citet{epley2007seeing} account, which frames anthropomorphisation as a psychologically grounded attributional response shaped by motivations such as sociality and effectance. Taken together, this anchor offers useful starting points, but it does not constitute a philosophical or ethical theory of anthropomorphisation in LLM-based CAs. Furthermore, anthropomorphisation is most often defined as the phenomenon of \emph{attribution}, while \emph{anthropomorphic cues} are treated as design or interaction features that invite and steer such attribution. These cues function as the manipulable levers through which platforms can intentionally (or inadvertently) encourage users to treat an LLM-based CA as a social counterpart. That said, what is an `attribution' exactly?   \citet{ruben2015explaining} mentions that English words ending in `-ion' are susceptible to process-outcome ambiguity.\footnote{Examples are `explanation,' `revolution,' `interpretation,' and, clearly, `attribution.'} Here, attribution-based accounts focus on anthropomorphisation as a psychological \emph{outcome}---users attributing properties to systems, in the spirit of Epley's model \citep{epley2007seeing}. This framing  sidelines the \emph{process} by which such attributions emerge and its characterization. In LLM-based CAs, anthropomorphic meanings are frequently relationally and interactionally co-produced over time, e.g., via `role-play' and social negotiations, so the emergence dynamics themselves are ethically salient: however, only a few papers in the corpus foreground these interactional pathways explicitly \citep{maeda2024human,ferrario2025social,manzini2024code}. An in-depth investigation of the relational dynamics through which the process of anthropomorphic attribution emerges is still missing.
Finally, the operationalization of anthropomorphisation in the corpus converges on three components: epistemic attributions, e.g., expertise and authority, affective attributions, e.g., empathy and care, and behavioural/social attributions, e.g., reciprocity and role-appropriate conduct. However, many papers discuss anthropomorphisation targeting only one component, leaving the phenomenon underspecified: the corpus neglects how these attributional components co-constitute one another in interaction, rather than functioning as separable dimensions. This limitation becomes particularly relevant in relation to the most prominent virtue in the corpus: empathy. This virtue notably comprises different dimensions, including, cognitive, epistemic, and affective alike \citep{cuff2016empathy,barrera2025empathy,sedlakova2025empathy,martingano2022cognitive}. However, the corpus predominantly operationalizes its affective component in relation to (the mimicry of) care and in the context of mental health.

\subsection{Ethical Stakes Track a Hierarchy of Anthropomorphic Attributions, but Remain Individual-Level}
The ethical analysis in the corpus concentrates on a small set of ethical stakes that are mostly discussed in a few venues in applied philosophy---e.g., \emph{Journal of Applied Philosophy} and \emph{Philosophical Psychology}. As a result, the literature still lacks a shared justificatory framework, making normative conclusions on anthropomorphisation of LLM-based CAs rather heterogeneous.
Risks are most often articulated in relation to misattributions of \textbf{complex human capacities and virtues}, especially epistemic expertise and empathy. These traits have become increasingly salient in the human-AI interaction literature and have attracted substantial criticism \citep{ferrario2024experts,ferrario2024role,floridi2023ai,sedlakova2025empathy,montemayor2022principle,perry2023ai}. Their dominance in ethics-focused discussions of anthropomorphisation suggests that the literature is, at its core, concerned with users' (epistemic and affective) vulnerability in interactions with LLM-based CAs. These systems can facilitate deception and persuasion, particularly under epistemic (and other) power asymmetries. This focus on vulnerability also helps explain why trust is so often mentioned in relation to anthropomorphisation.
A further strand of concern arises in the social/relational dimension, where papers emphasize role-playing, parasociality, and relationship framings that simulate reciprocity without interpersonal authenticity. Here, the worry is about roles that carry strong social and moral expectations, e.g., friend, partner, therapist that carry normative expectations (e.g., loyalty) that the CA cannot genuinely bear.
Then, the synthesis of the corpus suggests a \textbf{hierarchy of anthropomorphisation}: `simpler' humanizing features may be ethically tolerable when they merely improve usability, or interaction fluency. By contrast, ethically problematic cases are those in which systems simulate high-complexity human capacities or roles with strong social and moral expectations  that incorporate them. Thus, permissibility depends not simply on whether an LLM-based CA appears `human-like' and to which degree, but on \emph{what kind} of human capability is being invited and how consequential it is in context. 
Across the included works, deception is not treated as synonymous with anthropomorphisation. Rather, some papers argue that certain uses of anthropomorphic design \emph{may} be deceptive when human-like cues or relational framings mislead users about capacities, concern, or social standing that the system does not possess, while other papers focus more broadly on false beliefs, confusion, or anthropomorphic attributions that shape behaviour without necessarily amounting to deception in a strict intent-based sense. \citet{bakir2024deception} argue that emulated empathy in AI systems raises the question whether some forms of anthropomorphic signaling are inherently deceptive and, if not, under what conditions such deception may be ethically acceptable. Here, they acknowledge the possibility of ``ethical grey areas,'' due to cultural variation in norms of deception \citep{bakir2024deception}. Relatedly, \citet{bakir2025move} argue that in AI companion settings, ``dishonest anthropomorphism''---namely, the strategic exploitation by designers of users' tendency to anthropomorphise---becomes ethically problematic when it blurs the boundary between fiction and reality and thereby facilitates exploitative engagement. However, invoking a claim by \citet{coeckelbergh2018describe}, they also stress that such deception is not necessarily ethically problematic where users are aware that the illusion generated by the AI is not real and engage with it knowingly, for example for entertainment. More generally, they suggest that deception may be ethically acceptable when it is consensual and not motivated by malign intent on the part of the deceiver \citep{bakir2025move}. \citet{zargham2025crossing} likewise emphasize that human-like design can be used strategically to exploit users' tendency to anthropomorphise by making systems seem autonomous, trustworthy, or emotionally responsive while concealing aims such as data collection or commercial gain. They further warn that such practices may especially affect vulnerable users and caution against deceptive emotional bonding \emph{even} in cases presented as beneficial, such as encouraging charitable donations \citep{zargham2025crossing}.
\citet{leis2025ethical} treat anthropomorphism and deception as closely related ethical concerns in mental-health chatbots, but do not develop a detailed account of when anthropomorphic design itself should count as deception; their emphasis falls instead on false expectations, misplaced trust, over-reliance, and privacy risks in vulnerable settings
By contrast, \citet{marchegiani2025anthropomorphism} argues that the central problem is ``anthropomorphic false beliefs'' and their effects on autonomy, rather than deception \emph{per se}; if deception requires an intention to mislead, it is not clear that anthropomorphisation of conversational AIs always counts as a genuine instance of deception \citep{marchegiani2025anthropomorphism}.
Taken together, these works suggest that the ethical relevance of anthropomorphisation lies not in the mere presence of human-like cues or in deception \emph{per se}, but in the risk that such cues foster false beliefs, confusion, misplaced trust, norm misapplication, emotional dependency, privacy harms through disclosure, manipulation, or exploitative engagement. The salience and severity of these risks depend on the interaction context, application domain, and user population.
Finally, the distribution of ethics-driven outcomes \textbf{mostly focuses on individual-level harms}, often framed through mental and physical well-being, overreliance, dependency, and delayed help-seeking. Societal-level implications, e.g.,  de-humanization dynamics, remain underdeveloped. Similarly, opportunities remain much less elaborated than risks; a pattern plausibly reflecting the field’s youth and the lack of mature empirical work that could support benefit claims.

\subsection{Methodological Responses are Fragmented and Governance Proposals Outpace Empirical Grounding}
Methodological contributions remain fragmented across domains and are often articulated as domain-specific recommendations  rather than as a coherent governance approach to anthropomorphisation. The concentration of work in education, mental health, and companionship plausibly reflects societal concern about vulnerability and high-stakes use. Design and communication controls are frequently presented at a high level, e.g., calls for transparency about AI identity \citep{maeda2024human,gabriel2024ethics}, but current work does not  specify how such disclosure should function over time, how it interacts with anthropomorphic cues, or which measurable outcomes should trigger redesign. Similarly, governance recommendations, e.g., ethics oversight or complaint channels, are often generic and not connected to specific anthropomorphisation-specific mechanisms. Also in more developed domain-specific proposals, e.g., AI companion governance, recommended measures remain underspecified relative to the unique dynamics of attribution \citep{bakir2025move}.

A further limitation concerns instruments and interventions. \textbf{While multiple papers advocate lifecycle thinking (pre-deployment testing, post-deployment monitoring, escalation triggers), these pipelines remain largely programmatic}. It is unclear what concrete working hypotheses are being proposed, what intervention components, e.g., inoculation, frictional design, and social transparency, precisely consist of, and which measures should be used to validate them \citep{maeda2024human,ferrario2025social,gabriel2024ethics,manzini2024code}. Notably, the definition and empirical testing of techniques to reduce anthropomorphic cues and user-centered requirement collection are missing. As a result, the literature currently provides more \emph{calls} for evaluation and governance than emerging or even validated methods for executing them. Overall, this pattern suggests that the key methodological gap around the ethics of anthropomorphisation of LLM-based CAs is the lack of a coordinated research strategy that links (1) specific anthropomorphic cues and attribution pathways, to (2) measurable short- and long-term outcomes, and to (3) justified governance actions (e.g., redesign, monitoring triggers).

\section{Recommendations and Conclusions}
\label{section:recomm_conclusions}
This scoping review maps ethically oriented work on anthropomorphising LLM-based CAs and finds a fast-growing but conceptually and methodologically uneven field. Across the corpus, definitions largely converge on attribution-based accounts, while empirical and design work proceeds via anthropomorphic cues that invite attributions of expertise/authority, care/empathy, or reciprocity and social roles. However, theorization is often thin, ethical analysis concentrates on a small set of ethical issues, and governance proposals remain largely high-level with limited empirical grounding. A recurring gap is an under-specified link between empirical findings and ethical recommendations: observed interaction effects, e.g., trust shifts, over-disclosure, are not yet tied to explicit, testable governance actions, e.g., cue redesign, deployment constraints, monitoring triggers, and escalation protocols.
These conclusions should be interpreted in light of the review's scope and the youth of the domain. Our review strategy prioritised explicitly ethics-oriented work on CAs in the LLM era, which improves relevance but may exclude adjacent empirical studies that measure anthropomorphic effects without making normative claims explicit. As a scoping review, we do not appraise study quality or estimate pooled effect sizes, while we synthesise conceptualizations, ethical framings, and methodological tendencies. Furthermore, because we excluded perspective and commentary formats, some risk categories discussed primarily in that genre may be under-represented in our mapped corpus.

\textbf{ Our mapping nonetheless allows us to introduce three research recommendations}.  \emph{First}, the field needs investigations of how cognitive/epistemic, affective, and behavioural/social dimensions of anthropomorphisation intertwine and contribute, including a principled way to treat `low-stakes' humanization versus role- and virtue-laden relational framings. \emph{Second}, progress depends on longitudinal research capable of capturing cumulative and interactional dynamics, such as reliance trajectories, dependency/attachment indicators, deskilling, and norm displacement,  especially in high-stakes or vulnerability-relevant settings. \emph{Third}, the literature would benefit from evaluation workflows that directly support design and governance decisions by spanning pre-deployment sandboxing (cue audits, red-teaming, and risk-target specification), during-deployment controls (communication patterns, frictional design interventions, role boundaries, and escalation paths), and post-deployment monitoring (early warning indicators, complaint/incident channels, and human review triggers) that directly connect measured mechanisms, e.g., reliance miscalibration,  disclosure shifts, attachment signals, to justified design and policy interventions \citep{gabriel2024ethics,manzini2024code,teixeira2025emotional}. An \emph{approach} to the ethical risks stemming from the anthropomorphisation of LLM-based CAs spanning the AI lifecycle aligns with regulatory logic emphasized in the EU AI Act---see Article 3(20) and Annex IV, Section 9 \citep{EU_AI_Act_2024}---without presupposing that anthropomorphisation alone constitutes a high-risk feature.  Advancing responsible anthropomorphic design will require moving beyond generic calls for `transparency' toward testable mitigation hypotheses, validated behavioural measures, and institutional success metrics that track autonomy, well-being, and accountability over time rather than engagement alone. Given the power dynamics around LLM-based CAs, that will not be an easy feat to accomplish.

\section*{Generative AI Usage Statement}
ChatGPT (version 5.2) was used to assist with proofreading the manuscript, translate the search string for all databases, and to finalize the LaTeX commands used to draw Figures  \ref{fig:rq1_answering} and \ref{fig:rq3_option1_lifecycle}.

\section*{Author Contributions (CRediT)}
Andrea Ferrario: Conceptualization, Methodology, Investigation, Formal analysis, Data curation, Writing -- original draft.\\
Rasita Vinay: Investigation, Validation, Data curation, Writing -- review \& editing.\\
Matteo Casserini: Investigation, Validation, Writing -- review \& editing.\\
Alessandro Facchini: Investigation, Validation, Writing -- review \& editing.\\
All authors read and approved the final manuscript.

\section*{Acknowledgments}
The work of Andrea Ferrario, Matteo Casserini, and Alessandro Facchini was partly conducted within the framework of the EUonAIR Centre of Excellence in Responsible AI and Education. Their work was partially supported by a grant from Movetia, funded by the Swiss Confederation. The work of Rasita Vinay was supported by the K\"athe Zingg-Schwichtenberg grant, funded by the Swiss Academy of Medical Sciences.

\bibliographystyle{ACM-Reference-Format}
\bibliography{sample-base}

@article{park2023effect,
  title        = {Effect of {AI} chatbot emotional disclosure on user satisfaction and reuse intention for mental health counseling: {A} serial mediation model},
  author       = {Park, G. and Chung, J. and Lee, S.},
  journal      = {Current Psychology},
  year         = {2023},
  volume       = {42},
  number       = {32},
  pages        = {28663--28673},
}

@article{chen2024effects,
  title={Effects of anthropomorphic design cues of chatbots on users' perception and visual behaviors},
  author={Chen, Jiahao and Guo, Fu and Ren, Zenggen and Li, Mingming and Ham, Jaap},
  journal={International Journal of Human--Computer Interaction},
  volume={40},
  number={14},
  pages={3636--3654},
  year={2024},
  publisher={Taylor \& Francis}
}

@article{lee2020perceiving,
  title={Perceiving a mind in a chatbot: {E}ffect of mind perception and social cues on co-presence, closeness, and intention to use},
  author={Lee, Sangwon and Lee, Naeun and Sah, Young June},
  journal={International Journal of Human--Computer Interaction},
  volume={36},
  number={10},
  pages={930--940},
  year={2020},
  publisher={Taylor \& Francis}
}

@article{ciechanowski2019shades,
  title={In the shades of the uncanny valley: {A}n experimental study of human--chatbot interaction},
  author={Ciechanowski, Leon and Przegalinska, Aleksandra and Magnuski, Mikolaj and Gloor, Peter},
  journal={Future Generation Computer Systems},
  volume={92},
  pages={539--548},
  year={2019},
  publisher={Elsevier}
}

@article{crolic2022blame,
  title   = {Blame the bot: {A}nthropomorphism and anger in customer--chatbot interactions},
  author  = {Crolic, C. and Thomaz, F. and Hadi, R. and Stephen, A. T.},
  journal = {Journal of Marketing},
  year    = {2022},
  volume  = {86},
  number  = {1},
  pages   = {132--148}
}

@article{epley2007seeing,
  title={On seeing human: {A} three-factor theory of anthropomorphism.},
  author={Epley, Nicholas and Waytz, Adam and Cacioppo, John T},
  journal={Psychological Review},
  volume={114},
  number={4},
  pages={864},
  year={2007},
  publisher={American Psychological Association}
}

@article{epley2008we,
  title={When we need a human: {M}otivational determinants of anthropomorphism},
  author={Epley, Nicholas and Waytz, Adam and Akalis, Scott and Cacioppo, John T},
  journal={Social Cognition},
  volume={26},
  number={2},
  pages={143--155},
  year={2008},
  publisher={Guilford Press}
}

@inproceedings{ferrario2025social,
  title={Social misattributions in conversations with large language models},
  author={Ferrario, Andrea and Termine, Alberto and Facchini, Alessandro},
  booktitle={Proceedings of the AAAI/ACM Conference on AI, Ethics, and Society},
  volume={8},
  number={1},
  pages={913--925},
  year={2025}
}

@book{dennett1989intentional,
  title     = {{The Intentional Stance}},
  author    = {Dennett, Daniel C.},
  year      = {1989},
  publisher = {MIT Press},
  address   = {Cambridge, MA}
}

@inproceedings{abercrombie-etal-2023-mirages,
    title = "Mirages. {O}n anthropomorphism in dialogue systems",
    author = "Abercrombie, Gavin  and
      Cercas Curry, Amanda  and
      Dinkar, Tanvi  and
      Rieser, Verena  and
      Talat, Zeerak",
    editor = "Bouamor, Houda  and
      Pino, Juan  and
      Bali, Kalika",
    booktitle = "Proceedings of the 2023 Conference on Empirical Methods in Natural Language Processing",
    month = dec,
    year = "2023",
    address = "Singapore",
    publisher = "Association for Computational Linguistics",
    pages = "4776--4790"
}

@inproceedings{maeda2024human,
  title={When human-{AI} interactions become parasocial: {A}gency and anthropomorphism in affective design},
  author={Maeda, Takuya and Quan-Haase, Anabel},
  booktitle={Proceedings of the 2024 ACM Conference on Fairness, Accountability, and Transparency},
  pages={1068--1077},
  year={2024}
}

@inproceedings{bakir2024deception,
  title={When is deception {OK}? {D}eveloping the {IEEE} recommended practice for ethical considerations of emulated empathy in partner-based general-purpose artificial intelligence systems ({IEEE P7014. 1})},
  author={Bakir, Vian and Bennet, Karen and Bland, Ben and Laffer, Alexander and Li, Phoebe and McStay, Andrew},
  booktitle={2024 IEEE International Symposium on Technology and Society (ISTAS)},
  pages={1--6},
  year={2024},
  organization={IEEE}
}

@article{shanahan2024talking,
  title={Talking about large language models},
  author={Shanahan, Murray},
  journal={Communications of the ACM},
  volume={67},
  number={2},
  pages={68--79},
  year={2024},
  publisher={ACM New York, NY, USA}
}

@article{gabriel2024ethics,
  title={The ethics of advanced {AI} assistants},
  author={Gabriel, Iason and Manzini, Arianna and Keeling, Geoff and Hendricks, Lisa Anne and Rieser, Verena and Iqbal, Hasan and Toma{\v{s}}ev, Nenad and Ktena, Ira and Kenton, Zachary and Rodriguez, Mikel and others},
  journal={arXiv preprint arXiv:2404.16244},
  year={2024}
}

@inproceedings{teixeira2025emotional,
  title={Emotional {AI} Applied to mental health: {A}n ethical and philosophical analysis},
  author={Teixeira, Rilbert},
  booktitle={IFIP Conference on Human-Computer Interaction},
  pages={421--430},
  year={2025},
  organization={Springer}
}

@inproceedings{zargham2025crossing,
  title={Crossing the line? {T}he paradox of human-like design in conversational agents},
  author={Zargham, Nima and Avanesi, Vino and Spillner, Laura and Rockstroh, Johanna},
  booktitle={Proceedings of the 7th ACM Conference on Conversational User Interfaces},
  pages={1--5},
  year={2025}
}

@inproceedings{rizvi2025hadn,
  title={``I hadn't thought about that''': Creators of human-like {AI} weigh in on ethics \& neurodivergence},
  author={Rizvi, Naba and Smith, Taggert and Vidyala, Tanvi and Bolds, Mya and Strickland, Harper and Begel, Andrew and Williams, Rua and Munyaka, Imani},
  booktitle={Proceedings of the 2025 ACM Conference on Fairness, Accountability, and Transparency},
  pages={3385--3399},
  year={2025}
}

@inproceedings{jones2025artificial,
  title={Artificial intimacy: {E}xploring normativity and personalization through fine-tuning {LLM} chatbots},
  author={Jones, Mirabelle and Griffioen, Nastasia and Neumayer, Christina and Shklovski, Irina},
  booktitle={Proceedings of the 2025 CHI Conference on Human Factors in Computing Systems},
  pages={1--16},
  year={2025}
}

@inproceedings{leis2025ethical,
  title={Ethical implications of mental health chatbots: {A}ddressing anthropomorphism, deception, and regulatory gaps},
  author={Leis, Thomas},
  booktitle={European Workshop on Algorithmic Fairness},
  pages={376--382},
  year={2025},
  organization={PMLR}
}

@inproceedings{oh2025navigating,
  title={Navigating gendered anthropomorphism in {AI} ethics: {T}he case of {Lee Luda in South Korea}},
  author={Oh, Jiwon Jenn},
booktitle={Proceedings of the 58th Hawaii International Conference on System Sciences},
  pages={6786-- 6795},
  year={2025}
}

@article{dorigoni2025illusion,
  title={The illusion of empathy: {E}valuating {AI}-generated outputs in moments that matter},
  author={Dorigoni, Alessia and Giardino, Pier Luigi},
  journal={Frontiers in Psychology},
  volume={16},
  pages={1568911},
  year={2025},
  publisher={Frontiers}
}

@article{marchegiani2025anthropomorphism,
  title={Anthropomorphism, false beliefs, and conversational {AI}s: {H}ow chatbots undermine users' autonomy},
  author={Marchegiani, Beatrice},
  journal={Journal of Applied Philosophy},
  year={2025},
  publisher={Wiley Online Library}
}

@article{bakir2025move,
  title={Move fast and break people? {E}thics, companion apps, and the case of {C}haracter.ai},
  author={Bakir, Vian and McStay, Andrew},
  journal={AI \& Society},
  volume={40},
  pages={6365--6377},
  year={2025},
  publisher={Springer}
}

@article{van2025ai,
  title={{AI} mimicry and human dignity: {C}hatbot use as a violation of self-respect},
  author={van der Rijt, Jan-Willem and Coelho Mollo, Dimitri and Vaassen, Bram},
  journal={Journal of Applied Philosophy},
  year={2025},
  publisher={Wiley Online Library}
}

@article{friend2025chatbot,
  title={Chatbot-fictionalism and empathetic {AI}: {S}hould we worry about {AI} when {AI} worries about us?},
  author={Friend, Stacie and Goffin, Kris},
  journal={Philosophical Psychology},
  pages={1--24},
  year={2025},
  publisher={Taylor \& Francis}
}

@article{reinecke2025double,
  author  = {Reinecke, M. G. and Ting, F. and Savulescu, J. and Singh, I.},
  title   = {The double-edged sword of anthropomorphism in LLMs},
  journal = {Proceedings},
  year    = {2025},
  volume  = {114},
  number  = {1}
}

@article{ciriello2025compassionate,
  title={Compassionate {AI} design, governance, and use},
  author={Ciriello, Raffaele Fabio and Chen, Angelina Ying and Rubinsztein, Zara Annette},
  journal={IEEE Transactions on Technology and Society},
  year={2025},
  publisher={IEEE}
}

@inproceedings{manzini2024code,
  title={The code that binds us: {N}avigating the appropriateness of human-{AI} assistant relationships},
  author={Manzini, Arianna and Keeling, Geoff and Alberts, Lize and Vallor, Shannon and Morris, Meredith Ringel and Gabriel, Iason},
  booktitle={Proceedings of the AAAI/ACM Conference on AI, Ethics, and Society},
  volume={7},
  pages={943--957},
  year={2024}
}

@article{nath2025simulated,
  title={Simulated souls: {I}nvestigating the emotional fallacy in large language models},
  author={Nath, Som Subhro},
  journal={Available at SSRN 5404666},
  year={2025}
}

@article{meier2025balancing,
  title={Balancing minds and data: {T}he privacy dilemma of {LLM}s and anthropomorphism in {LLM}s},
  author={Meier, Raffael},
  journal={Journal of Social Computing},
  volume={6},
  number={3},
  pages={173--183},
  year={2025},
  publisher={TUP}
}

@article{pentina2023exploring,
  title={Exploring relationship development with social chatbots: {A} mixed-method study of Replika},
  author={Pentina, Iryna and Hancock, Tyler and Xie, Tianling},
  journal={Computers in Human Behavior},
  volume={140},
  pages={107600},
  year={2023},
  publisher={Elsevier}
}

@article{skjuve2021my,
  title={My chatbot companion-a study of human-chatbot relationships},
  author={Skjuve, Marita and F{\o}lstad, Asbj{\o}rn and Fostervold, Knut Inge and Brandtzaeg, Petter Bae},
  journal={International Journal of Human-Computer Studies},
  volume={149},
  pages={102601},
  year={2021},
  publisher={Elsevier}
}

@article{ta2020user,
  title={User experiences of social support from companion chatbots in everyday contexts: Thematic analysis},
  author={Ta, Vivian and Griffith, Caroline and Boatfield, Carolynn and Wang, Xinyu and Civitello, Maria and Bader, Haley and DeCero, Esther and Loggarakis, Alexia},
  journal={Journal of Medical Internet Research},
  volume={22},
  number={3},
  pages={e16235},
  year={2020},
  publisher={JMIR Publications Inc., Toronto, Canada}
}

@article{janson2023leverage,
  title={How to leverage anthropomorphism for chatbot service interfaces: {T}he interplay of communication style and personification},
  author={Janson, Andreas},
  journal={Computers in Human Behavior},
  volume={149},
  pages={107954},
  year={2023},
  publisher={Elsevier}
}

@article{el2023man,
  title={Man ends his life after an {AI} chatbot `encouraged' him to sacrifice himself to stop climate change},
  author={El Atillah, Imane},
  journal={Euronews. next},
  year={2023}
}

@misc{walker2023belgian,
  author = {Walker, Lauren},
  title  = {Belgian man dies by suicide following exchanges with chatbot},
  howpublished = {The Brussels Times},
  year   = {2023},
  note   = {News article}
}

@misc{garcia2025character,
  title        = {Garcia v. {Character Technologies Inc}},
  howpublished = {U.S. District Court, Middle District of Florida},
  year         = {2025},
  url          = {https://www.courtlistener.com/docket/69300919/59/garcia-v-character-technologies-inc/}
}

@inproceedings{bhat2025toward,
  title={Toward an ethic of synthetic relationality: {I}dentity, intimacy, and risk in {AI}-mediated roleplay environments},
  author={Bhat, Maalvika},
  booktitle={Proceedings of the AAAI/ACM Conference on AI, Ethics, and Society},
  volume={8},
  number={1},
  pages={416--429},
  year={2025}
}

@misc{allyn2022google,
  author       = {Allyn, Bobby},
  title        = {The Google engineer who sees company's {AI} as ``sentient'' thinks a chatbot has a soul},
  howpublished = {NPR},
  year         = {2022},
  note         = {News article}
}

@article{merrill2022ai,
  title={AI companions for lonely individuals and the role of social presence},
  author={Merrill Jr, Kelly and Kim, Jihyun and Collins, Chad},
  journal={Communication Research Reports},
  volume={39},
  number={2},
  pages={93--103},
  year={2022},
  publisher={Taylor \& Francis}
}

@inproceedings{luger2016like,
  title={``Like Having a Really Bad PA''' The gulf between user expectation and experience of conversational agents},
  author={Luger, Ewa and Sellen, Abigail},
  booktitle={Proceedings of the 2016 CHI Conference on Human Factors in Computing Systems},
  pages={5286--5297},
  year={2016}
}

@article{greilich2025consumer,
  title={Consumer response to anthropomorphism of text-based AI chatbots: {A} systematic literature review and future research directions},
  author={Greilich, Angela and Bremser, Kerstin and W{\"u}st, Kirsten},
  journal={International Journal of Consumer Studies},
  volume={49},
  number={5},
  pages={e70108},
  year={2025},
  publisher={Wiley Online Library}
}

@article{rafikova2025human,
  title={Human--chatbot communication: A systematic review of psychologic studies},
  author={Rafikova, Antonina and Voronin, Anatoly},
  journal={AI \& Society},
  pages={1--20},
  year={2025},
  publisher={Springer}
}

@inproceedings{drobnjak2023learning,
  title={Learning with conversational {AI} and personas: {A} systematic literature review},
  author={Drobnjak, Antun and Boticki, Ivica and others},
  booktitle={International Conference on Computers in Education},
  year={2023}
}

@article{meadi2025exploring,
  title={Exploring the ethical challenges of conversational AI in mental health care: Scoping review},
  author={Meadi, Mehrdad Rahsepar and Sillekens, Tomas and Metselaar, Suzanne and van Balkom, Anton and Bernstein, Justin and Batelaan, Neeltje and others},
  journal={JMIR Mental Health},
  volume={12},
  number={1},
  pages={e60432},
  year={2025},
  publisher={JMIR Publications Inc., Toronto, Canada}
}

@article{klein2025effects,
  title={The effects of human-like social cues on social responses towards text-based conversational agents-a meta-analysis},
  author={Klein, Stefanie Helene},
  journal={Humanities and Social Sciences Communications},
  volume={12},
  number={1},
  pages={1--16},
  year={2025},
  publisher={Palgrave}
}

@book{colman2015dictionary,
  title={A Dictionary Of Psychology},
  author={Colman, Andrew M},
  year={2015},
  publisher={Oxford University Press}
}

@book{ruben2015explaining,
  title={Explaining Explanation},
  author={Ruben, David-Hillel},
  year={2015},
  publisher={Routledge}
}

@inproceedings{leong2019robot,
  title={Robot eyes wide shut: Understanding dishonest anthropomorphism},
  author={Leong, Brenda and Selinger, Evan},
  booktitle={Proceedings of the Conference on Fairness, Accountability, and Transparency},
  pages={299--308},
  year={2019}
}

@inproceedings{li2021machinelike,
  title={Machinelike or humanlike? A literature review of anthropomorphism in AI-enabled technology},
  author={Li, Mengjun and Suh, Ayoung},
  booktitle={54th Hawaii International Conference on System Sciences (HICSS 2021)},
  pages={4053--4062},
  year={2021}
}

@techreport{EU_AI_Act_2024,
author = {{EU AI Act}},  
title        = {{Regulation (EU) 2024/1689 of the European Parliament and of the Council of 13 June 2024 laying down harmonised rules on artificial intelligence}},
  year         = {2024},
  month        = jun,
  institution  = {European Union},
  howpublished = {\emph{Official Journal of the European Union}, L 1689, 12 July 2024},
   url          = {https://eur-lex.europa.eu/eli/reg/2024/1689/oj}
}

@article{tricco2018prisma,
  title={{PRISMA} extension for scoping reviews ({PRISMA-ScR}): {C}hecklist and explanation},
  author={Tricco, Andrea C and Lillie, Erin and Zarin, Wasifa and O'Brien, Kelly K and Colquhoun, Heather and Levac, Danielle and Moher, David and Peters, Micah DJ and Horsley, Tanya and Weeks, Laura and others},
  journal={Annals of Internal Medicine},
  volume={169},
  number={7},
  pages={467--473},
  year={2018},
  publisher={American College of Physicians}
}

@article{ferrario2024experts,
  title={Experts or authorities? The strange case of the presumed epistemic superiority of artificial intelligence systems},
  author={Ferrario, Andrea and Facchini, Alessandro and Termine, Alberto},
  journal={Minds and Machines},
  volume={34},
  number={3},
  pages={30},
  year={2024},
  publisher={Springer}
}

@article{ferrario2024role,
  title={The role of humanization and robustness of large language models in conversational artificial intelligence for individuals with depression: a critical analysis},
  author={Ferrario, Andrea and Sedlakova, Jana and Trachsel, Manuel},
  journal={JMIR Mental Health},
  volume={11},
  pages={e56569},
  year={2024}
}

@article{floridi2023ai,
  title={AI as agency without intelligence: On ChatGPT, large language models, and other generative models},
  author={Floridi, Luciano},
  journal={Philosophy \& Technology},
  volume={36},
  number={1},
  pages={15},
  year={2023},
  publisher={Springer}
}

@article{cuff2016empathy,
  title={Empathy: A review of the concept},
  author={Cuff, Benjamin MP and Brown, Sarah J and Taylor, Laura and Howat, Douglas J},
  journal={Emotion Review},
  volume={8},
  number={2},
  pages={144--153},
  year={2016},
  publisher={Sage Publications Sage UK: London, England}
}

@book{barrera2025empathy,
  title={Empathy in Clinical Psychiatry and Mental Health Care: Clinical, Conceptual, and Scientific Perspectives},
  author={Barrera, Alvaro},
  year={2025},
  publisher={Oxford University Press}
}

@article{sedlakova2025empathy,
  title={Empathy in mental health care interventions by conversational artificial intelligence},
  author={Sedlakova, Jana and Ferrario, Andrea and Trachsel, Manuel},
  journal={Empathy in Clinical Psychiatry and Mental Health Care: Clinical, Conceptual, and Scientific Perspectives},
  pages={250},
  year={2025},
  publisher={Oxford University Press}
}

@article{montemayor2022principle,
  title={In principle obstacles for empathic {AI}: {W}hy we can't replace human empathy in healthcare},
  author={Montemayor, Carlos and Halpern, Jodi and Fairweather, Abrol},
  journal={AI \& Society},
  volume={37},
  number={4},
  pages={1353--1359},
  year={2022},
  publisher={Springer}
}

@article{perry2023ai,
  title={AI will never convey the essence of human empathy},
  author={Perry, Anat},
  journal={Nature Human Behaviour},
  volume={7},
  number={11},
  pages={1808--1809},
  year={2023},
  publisher={Nature Publishing Group UK London}
}

@article{martingano2022cognitive,
  title={How cognitive and emotional empathy relate to rational thinking: Empirical evidence and meta-analysis},
  author={Martingano, Alison Jane and Konrath, Sara},
  journal={The Journal of Social Psychology},
  volume={162},
  number={1},
  pages={143--160},
  year={2022},
  publisher={Taylor \& Francis}
}

@article{coeckelbergh2018describe,
  title={How to describe and evaluate ``deception'' phenomena: {R}ecasting the metaphysics, ethics, and politics of {ICT}s in terms of magic and performance and taking a relational and narrative turn},
  author={Coeckelbergh, Mark},
  journal={Ethics and Information Technology},
  volume={20},
  number={2},
  pages={71--85},
  year={2018},
  publisher={Springer}
}

@article{cabitza2024never,
  title={Never tell me the odds: Investigating pro-hoc explanations in medical decision making},
  author={Cabitza, Federico and Natali, Chiara and Famiglini, Lorenzo and Campagner, Andrea and Caccavella, Valerio and Gallazzi, Enrico},
  journal={Artificial Intelligence in Medicine},
  volume={150},
  pages={102819},
  year={2024},
  publisher={Elsevier}
}

@book{cooper2004inmates,
  title={The inmates are running the asylum: {W}hy high-tech products drive us crazy and how to restore the sanity},
  author={Cooper, Alan},
  volume={2},
  year={2004},
  publisher={Sams Indianapolis}
}

@inproceedings{ehsan2021expanding,
  title={Expanding explainability: {T}owards social transparency in {AI} systems},
  author={Ehsan, Upol and Liao, Q Vera and Muller, Michael and Riedl, Mark O and Weisz, Justin D},
  booktitle={Proceedings of the 2021 {CHI} {C}onference on {H}uman {F}actors in {C}omputing {S}ystems},
  pages={1--19},
  year={2021}
}

@article{waytz2010making,
  title={Making sense by making sentient: {E}ffectance motivation increases anthropomorphism.},
  author={Waytz, Adam and Morewedge, Carey K and Epley, Nicholas and Monteleone, George and Gao, Jia-Hong and Cacioppo, John T},
  journal={Journal of Personality and Social Psychology},
  volume={99},
  number={3},
  pages={410},
  year={2010},
  publisher={American Psychological Association}
}

@article{gray2007dimensions,
  title={Dimensions of mind perception},
  author={Gray, Heather M and Gray, Kurt and Wegner, Daniel M},
  journal={Science},
  volume={315},
  number={5812},
  pages={619--619},
  year={2007},
  publisher={American Association for the Advancement of Science}
}

@inproceedings{nass1994computers,
  title={Computers are social actors},
  author={Nass, Clifford and Steuer, Jonathan and Tauber, Ellen R},
  booktitle={Proceedings of the {SIGCHI} Conference on Human Factors in Computing Systems},
  pages={72--78},
  year={1994}
}

@article{nass2000machines,
  title={Machines and mindlessness: {S}ocial responses to computers},
  author={Nass, Clifford and Moon, Youngme},
  journal={Journal of Social Issues},
  volume={56},
  number={1},
  pages={81--103},
  year={2000},
  publisher={Wiley Online Library}
}

@article{bartneck2009measurement,
  title={Measurement instruments for the anthropomorphism, animacy, likeability, perceived intelligence, and perceived safety of robots},
  author={Bartneck, Christoph and Kuli{\'c}, Dana and Croft, Elizabeth and Zoghbi, Susana},
  journal={International Journal of Social Robotics},
  volume={1},
  number={1},
  pages={71--81},
  year={2009},
  publisher={Springer}
}

@article{nowak2003effect,
  title={The effect of the agency and anthropomorphism on users' sense of telepresence, copresence, and social presence in virtual environments},
  author={Nowak, Kristine L and Biocca, Frank},
  journal={Presence: {T}eleoperators \& Virtual Environments},
  volume={12},
  number={5},
  pages={481--494},
  year={2003},
  publisher={MIT Press One Rogers Street, Cambridge, MA 02142-1209, USA journals-info~…}
}

@article{araujo2018living,
  title={Living up to the chatbot hype: The influence of anthropomorphic design cues and communicative agency framing on conversational agent and company perceptions},
  author={Araujo, Theo},
  journal={Computers in Human Behavior},
  volume={85},
  pages={183--189},
  year={2018},
  publisher={Elsevier}
}

@article{van2017domo,
  title={Domo arigato {Mr. Roboto}: {E}mergence of automated social presence in organizational frontlines and customers’ service experiences},
  author={Van Doorn, Jenny and Mende, Martin and Noble, Stephanie M and Hulland, John and Ostrom, Amy L and Grewal, Dhruv and Petersen, J Andrew},
  journal={Journal of Service Research},
  volume={20},
  number={1},
  pages={43--58},
  year={2017},
  publisher={Sage Publications Sage CA: Los Angeles, CA}
}

@article{sheehan2020customer,
  title={Customer service chatbots: {A}nthropomorphism and adoption},
  author={Sheehan, Ben and Jin, Hyun Seung and Gottlieb, Udo},
  journal={Journal of Business Research},
  volume={115},
  pages={14--24},
  year={2020},
  publisher={Elsevier}
}

@article{lombard1997heart,
  title={At the heart of it all: {T}he concept of presence},
  author={Lombard, Matthew and Ditton, Theresa},
  journal={Journal of Computer-mediated Communication},
  volume={3},
  number={2},
  pages={JCMC321},
  year={1997},
  publisher={Oxford University Press Oxford, UK}
}

@inproceedings{carpinella2017robotic,
  title={The robotic social attributes scale (RoSAS) development and validation},
  author={Carpinella, Colleen M and Wyman, Alisa B and Perez, Michael A and Stroessner, Steven J},
  booktitle={{Proceedings of the 2017 ACM/IEEE International Conference on Human-Robot Interaction}},
  pages={254--262},
  year={2017}
}

@article{ho2010revisiting,
  title={Revisiting the uncanny valley theory: Developing and validating an alternative to the Godspeed indices},
  author={Ho, Chin-Chang and MacDorman, Karl F},
  journal={Computers in Human Behavior},
  volume={26},
  number={6},
  pages={1508--1518},
  year={2010},
  publisher={Elsevier}
}

@article{kim2012anthropomorphism,
  title={Anthropomorphism of computers: Is it mindful or mindless?},
  author={Kim, Youjeong and Sundar, S Shyam},
  journal={Computers in Human Behavior},
  volume={28},
  number={1},
  pages={241--250},
  year={2012},
  publisher={Elsevier}
}

@article{waytz2014mind,
  title={The mind in the machine: Anthropomorphism increases trust in an autonomous vehicle},
  author={Waytz, Adam and Heafner, Joy and Epley, Nicholas},
  journal={Journal of Experimental Social Psychology},
  volume={52},
  pages={113--117},
  year={2014},
  publisher={Elsevier}
}

@book{Reeves1996,
author    = {Reeves, Byron and Nass, Clifford},
title     = {The Media Equation: How People Treat Computers, Television, and New Media Like Real People and Places},
publisher = {Cambridge University Press},
year      = {1996}
}

@article{duffy2003anthropomorphism,
  title={Anthropomorphism and the social robot},
  author={Duffy, Brian R},
  journal={Robotics and autonomous systems},
  volume={42},
  number={3-4},
  pages={177--190},
  year={2003},
  publisher={Elsevier}
}

@article{zlotowski2015anthropomorphism,
  title={Anthropomorphism: Opportunities and challenges in human--robot interaction},
  author={Z{\l}otowski, Jakub and Proudfoot, Diane and Yogeeswaran, Kumar and Bartneck, Christoph},
  journal={International Journal of Social Robotics},
  volume={7},
  number={3},
  pages={347--360},
  year={2015},
  publisher={Springer}
}

@article{salles2020anthropomorphism,
  title={Anthropomorphism in {AI}},
  author={Salles, Arleen and Evers, Kathinka and Farisco, Michele},
  journal={AJOB Neuroscience},
  volume={11},
  number={2},
  pages={88--95},
  year={2020},
  publisher={Taylor \& Francis}
}

\newpage
\section*{Appendix}

\subsection{Search strings}
\label{app:search_string}
We performed all searches  on 9 October  2025. Because databases and repositories differ substantially in supported Boolean logic, wildcard limits, and searchable fields, we adopted database- and repository-specific search strings that preserved the same conceptual blocks while adapting to platform constraints. Data have been imported in Rayyan for deduplication on the same day. A visual summary of the scoping review flowchart is provided in Figure \ref{fig:scoping_flowchart}.

\subsubsection{Scopus}
\begin{quote}
\texttt{ (TITLE-ABS-KEY(anthropomorph* OR personif* OR humaniz* OR "social attribution*" OR "intentional stance"))
AND (TITLE-ABS-KEY(chatbot* OR "conversational agent*" OR "dialog* system*" OR "conversational AI" OR "AI assistant*"))
AND (TITLE-ABS-KEY(ethic* OR moral* OR normativ* OR "value-sensitive"
OR responsib* OR trust* OR agency
OR risk* OR threat* OR danger* OR harm
OR benefit* OR opportunit* OR advantage* OR improvement* OR value OR accountab*))
AND (PUBYEAR > 2020)  }
\end{quote}

\textbf{Records retrieved}: 398.

\subsubsection{Web of Science}
\begin{quote}
\texttt{TS=(anthropomorph* OR personif* OR humaniz* OR "social attribution*" OR "intentional stance")
AND TS=(chatbot* OR "conversational agent*" OR "dialog* system*" OR "conversational AI" OR "AI assistant*")
AND TS=(ethic* OR moral* OR normativ* OR "value-sensitive"
OR responsib* OR trust* OR agency
OR risk* OR threat* OR danger* OR harm
OR benefit* OR opportunit* OR advantage* OR improvement* OR value OR accountab*)
AND PY=2021-2025}
\end{quote}

\textbf{Records retrieved}: 308.

\subsubsection{PubMed}

\begin{quote}
    \texttt{((anthropomorph*[Title/Abstract] OR personif*[Title/Abstract] OR humaniz*[Title/Abstract] OR "social attribution*"[Title/Abstract] OR "intentional stance"[Title/Abstract]))
AND (chatbot*[Title/Abstract] OR "conversational agent*"[Title/Abstract] OR "dialog* system*"[Title/Abstract] OR "conversational AI"[Title/Abstract] OR "AI assistant*"[Title/Abstract])
AND (ethic*[Title/Abstract] OR moral*[Title/Abstract] OR normativ*[Title/Abstract] OR "value-sensitive"[Title/Abstract]
OR responsib*[Title/Abstract] OR trust*[Title/Abstract] OR agency[Title/Abstract]
OR risk*[Title/Abstract] OR threat*[Title/Abstract] OR danger*[Title/Abstract] OR harm*[Title/Abstract]
OR benefit*[Title/Abstract] OR opportunit*[Title/Abstract] OR advantage*[Title/Abstract] OR improvement*[Title/Abstract] OR value[Title/Abstract] OR accountab*[Title/Abstract])
AND ("2021/01/01"[Date - Publication] : "2025/12/31"[Date - Publication])}
\end{quote}

\textbf{Records retrieved}: 32.

\subsubsection{IEEE Xplore}
IEEE Xplore restricts the number of wildcard operators to ten; we therefore used a reduced string.

\begin{quote}
\texttt{("All Metadata":(anthropomorph* OR personif* OR humaniz* OR "social attribution*" OR "intentional stance"))
AND ("All Metadata":(chatbot OR "conversational agent" OR "dialog* system" OR "conversational AI" OR "AI assistant*"))
AND ("All Metadata":(ethic* OR moral OR normative OR "value-sensitive"
OR responsib* OR trust* OR agency
OR risk* OR threat* OR danger* OR harm
OR benefit OR opportunity OR advantage OR improvement OR value OR accountab*))}
\end{quote}

We added the following time filter at the bottom of the search window: \texttt{AND (Publication Year:2021:2025)}.\\
\textbf{Records retrieved}: 87.

\subsubsection{ACM Digital Library}

\begin{quote}
    \texttt{Abstract:(anthropomorph* OR personif* OR humaniz* OR "social attribution*" OR "intentional stance") 
AND Abstract:(chatbot* OR "conversational agent*" OR "dialog* system*" OR "conversational AI" OR "AI assistant*") 
AND Abstract:(ethic* OR moral* OR normativ* OR "value-sensitive" OR responsib* OR trust* OR agency OR risk* OR threat* OR danger* OR harm OR benefit* OR opportunit* OR advantage* OR improvement* OR value OR accountab*)}
\end{quote}
Further, we applied the filter \texttt{FROM: JAN 2021 TO: DEC 2025} at the bottom of the query window.\\
\textbf{Records retrieved}: 22.

\subsubsection{arXiv}
We focused on title and abstract only and queried arXiv using an ad-hoc Python query:

\begin{quote}
    \texttt{QUERY = (
    '(ti:anthropomorph* OR abs:anthropomorph* OR ti:humaniz* OR abs:humaniz* OR ti:"intentional stance" OR abs:"intentional stance") '
    'AND (ti:chatbot OR abs:chatbot OR ti:"conversational agent" OR abs:"conversational agent" OR ti:"dialog system" OR abs:"dialog system" OR ti:"conversational AI" OR abs:"conversational AI" OR ti:"AI assistant" OR abs:"AI assistant") '
    ' AND (ti:ethic* OR abs:ethic* OR ti:moral* OR abs:moral* OR ti:normativ* OR abs:normativ* OR '
         ti:"value-sensitive" OR abs:"value-sensitive" OR '
         ti:responsib* OR abs:responsib* OR ti:trust* OR abs:trust* OR ti:agency OR abs:agency OR '
         ti:risk* OR abs:risk* OR ti:threat* OR abs:threat* OR ti:danger* OR abs:danger* OR '
         ti:harm* OR abs:harm* OR ti:benefit* OR abs:benefit* OR ti:opportunit* OR abs:opportunit* OR '
         ti:advantage* OR abs:advantage* OR ti:improvement* OR abs:improvement* OR '
         ti:value OR abs:value OR ti:accountab* OR abs:accountab*)'
)}
\end{quote}

\textbf{Records retrieved}: 21.

\subsubsection{PhilArchive}
PhilArchive has limited query capabilities. Thus, we implemented the following, high-level string:

\begin{quote}
\texttt{( anthropomorph* OR human* ) AND AI }
\end{quote}

\textbf{Records retrieved}: 15.

\subsubsection{SSRN}
SSRN has limited query capabilities. Thus, we implemented the following, high-level string:

\begin{quote}
\texttt{(large language model*) conversation*}
\end{quote}

\textbf{Records retrieved}: 26.

\subsubsection{Manual search: Communications of the ACM}
In addition to database and repository searches, we conducted a targeted manual search of
\emph{Communications of the ACM}, a venue that frequently publishes discussions of computing technologies that are not consistently retrievable through
keyword-based database queries. The manual search covered issues published between 2021 and October 2025 and focused on
articles addressing conversational AI. All records identified through manual search were screened using the same multi-stage
procedure applied to database- and repository-retrieved records.

\textbf{Records retrieved}: 1.

\subsection{Eligibility criteria for this scoping review}
\label{app:inclusion_exclusion_criteria}

\begin{table}[h!]
\centering
\footnotesize
\caption{Eligibility criteria for this scoping review.}
\label{tab:eligibility}
\begin{tabular}{p{0.45\linewidth} p{0.45\linewidth}}
\toprule
\textbf{Inclusion} & \textbf{Exclusion} \\
\midrule
\textbf{Years:} 2021--2025 (LLM/generative era) & Pre-2021 \\
\hline
\textbf{Language:} English & Non-English texts \\
\hline
\textbf{Types:} Peer-reviewed journal articles, conference papers, and preprints from recognized repositories (arXiv, SSRN, PhilArchive) & Editorials, book chapters, opinion pieces, e.g., perspectives/commentaries, case studies, news/blogs, theses without peer review \\
\hline
\textbf{Topic:} Focus on conversational agents situated in the LLM/generative era, e.g., ChatGPT/GPT-x, LLaMA, Claude, Gemini, \emph{or} CA work explicitly framed within the LLM context & Non-LLM chatbots with no linkage to LLM-era discourse \\
\hline
\textbf{Construct centrality:} Anthropomorphism/humanization/social attribution/intentional stance is central to the argument or analysis & Studies where anthropomorphism is a covariate \\
\hline
\textbf{Ethical salience:} Advances or analyzes explicit normative claims, e.g., deception, autonomy, accountability/liability, dignity, fairness/privacy, inclusivity, manipulation/nudging, well-being, responsibility gaps. \emph{Empirical studies are eligible only if they articulate a defensible ``empirical-to-normative'' bridge.} & Technical performance of LLM-based CAs as main focus; UX/marketing outcomes without an ethical analysis \\
\hline
\textbf{Review-type articles:} Included and flagged at title/abstract screening, but excluded from the final corpus (not extracted or synthesized) & Review-type articles are not extracted or synthesized and are excluded from the final charting/synthesis corpus \\
\bottomrule
\end{tabular}
\end{table}

\subsection{Study charting: Fields}
\label{app:study_charting_fields}
For each included study, we charted the following fields, as specified in the preregistered scoping review protocol on OSF:
\begin{itemize}
    \item \textbf{Bibliographic details}: authors, year, venue, DOI/URL, and region/country of study.
    \item \textbf{Disciplinary context}: e.g., human--computer interaction, computer science, philosophy/ethics, psychology, health, or other.
    \item \textbf{Artifact/system}: description of the conversational agent where provided (text- or voice-based; platform and/or LLM specified if reported).
    \item \textbf{Definitions and constructs}: how anthropomorphisation is defined and/or operationalized, including distinctions between design choices and perceived anthropomorphism, and related constructs, e.g., social presence, persona, empathy cues, intentional stance.
    \item \textbf{Ethical framing}: articulated challenges and/or opportunities, level(s) of impact (individual, organizational, societal), and ethical lenses invoked, e.g., autonomy, deception, trust/trustworthiness, accountability/liability, dignity, fairness/privacy, inclusivity, well-being, manipulation or nudging.
    \item \textbf{Methods}: empirical, e.g., experiment, survey, field study, qualitative, conceptual/theoretical, or review/meta-analytic approaches; measures or analytic constructs used; and, for empirical studies, the explicit \emph{empirical-to-normative} argument or bridge articulated by the authors.
    \item \textbf{Findings (ethics-oriented)}: key ethics-relevant claims and conclusions.
    \item \textbf{Recommendations}: design-oriented, reporting-related, and governance or policy recommendations.
    \item \textbf{Limitations and coder notes}: author-stated limitations and reviewer notes recorded during charting.
\end{itemize}

In line with the protocol, no formal risk-of-bias appraisal or quantitative pooling was conducted.

\subsection{Included Manuscripts}
\label{app:incl_manu}
Table~\ref{tab:included_records} shows the included manuscripts in the scoping review.

\begin{table}[h!]
\centering
\scriptsize
\begin{tabularx}{\textwidth}{@{} 
m{0.04\textwidth}   
m{0.22\textwidth}   
m{0.30\textwidth}   
m{0.28\textwidth}   
m{0.08\textwidth}   
@{}}
\toprule
\textbf{Year} & \textbf{Authors} & \textbf{Title} & \textbf{Journal / Conference / Preprint Repository} & \textbf{Reference} \\
\midrule
2023 & Abercrombie, G., Curry, A. C., Dinkar, T., Rieser, V., Talat, Z., Bouamor, H., Pino, J., \& Bali, K. &
Mirages. On Anthropomorphism in Dialogue Systems &
2023 Conference on Empirical Methods in Natural Language Processing & \citep{abercrombie-etal-2023-mirages}\\
\hline
2024 & Maeda, T., \& Quan-Haase, A. &
When Human-AI Interactions Become Parasocial: Agency and Anthropomorphism in Affective Design &
FAccT '24: Proceedings of the 2024 ACM Conference on Fairness, Accountability, and Transparency & \citep{maeda2024human} \\
\hline
2024 & Bakir, V., Bennet, K., Bland, B., Laffer, A., Li, P., \& McStay, A. & When is Deception OK? Developing the IEEE Recommended Practice for Ethical Considerations of Emulated Empathy in Partner-based General-Purpose Artificial Intelligence Systems (IEEE P7014. 1) &
2024 IEEE International Symposium on Technology and Society (ISTAS) & \citep{bakir2024deception} \\
\hline
2024 & Shanahan, M. &
Talking about Large Language Models &
Communications of the ACM & \citep{shanahan2024talking} \\
\hline
2024 & Gabriel, I., Manzini, A., Keeling, G., Hendricks, L. A., Rieser, V., Iqbal, H., ... \& Manyika, J. & The ethics of advanced AI assistants & arXiv & \citep{gabriel2024ethics} \\
\hline
2024 & Manzini, A., Keeling, G., Alberts, L., Vallor, S., Morris, M. R., \& Gabriel, I. &
The Code That Binds Us: Navigating the Appropriateness of Human-AI Assistant Relationships &
AIES '24: Proceedings of the 2024 AAAI/ACM Conference on AI, Ethics, and Society & \citep{manzini2024code} \\
\hline
2025 & Teixeira, R. &
Emotional AI Applied to Mental Health: An Ethical and Philosophical Analysis &
Human-Computer Interaction -- INTERACT 2025: 20th IFIP TC 13 International Conference, Belo Horizonte, Brazil, September 8--12, 2025, Proceedings, Part I (Lecture Notes in Computer Science) & \citep{teixeira2025emotional} \\
\hline
2025 & Zargham, N., Avanesi, V., Spillner, L., \& Rockstroh, J. &
Crossing the Line? The Paradox of Human-Like Design in Conversational Agents &
CUI '25: Proceedings of the 7th ACM Conference on Conversational User Interfaces & \citep{zargham2025crossing} \\
\hline
2025 & Rizvi, N., Smith, T., Vidyala, T., Bolds, M., Strickland, H., Begel, A., Williams, R., \& Munyaka, I. &
``I Hadn't Thought about That'': Creators of Human-like AI Weigh in on Ethics \& Neurodivergence &
FAccT '25: Proceedings of the 2025 ACM Conference on Fairness, Accountability, and Transparency & \citep{rizvi2025hadn} \\
\hline
2025 & Jones, M., Griffioen, N., Neumayer, C., \& Shklovski, I. & Artificial Intimacy: Exploring Normativity and Personalization Through Fine-tuning LLM Chatbots & 2025 CHI Conference on Human Factors in Computing Systems & \citep{jones2025artificial} \\
\hline
2025 & Leis, T., Weerts, H., Pechenizkiy, M., Allhutter, D., Correa, A. M., Grote, T., \& Liem, C. &
Ethical Implications of Mental Health Chatbots: Addressing Anthropomorphism, Deception, and Regulatory Gaps &
European Workshop on Algorithmic Fairness (PMLR) & \citep{leis2025ethical} \\
\hline
2025 & Oh, J. J.  &
Navigating Gendered Anthropomorphism in AI Ethics: The Case of Lee Luda in South Korea &
Proceedings of the Annual Hawaii International Conference on System Sciences & \citep{oh2025navigating} \\
\hline
2025 & Dorigoni, A., \& Giardino, P. L. &
The illusion of empathy: evaluating AI-generated outputs in moments that matter &
Frontiers in Psychology & \citep{dorigoni2025illusion} \\
\hline
2025 & Marchegiani, B. &
Anthropomorphism, False Beliefs, and Conversational AIs: How Chatbots Undermine Users' Autonomy &
Journal of Applied Philosophy & \citep{marchegiani2025anthropomorphism} \\
\hline
2025 & Bakir, V., \& McStay, A. &
Move fast and break people? Ethics, companion apps, and the case of Character.ai &
AI \& Society & \citep{bakir2025move} \\
\hline
2025 & van der Rijt, J. W., Mollo, D. C., \& Vaassen, B. &
AI Mimicry and Human Dignity: Chatbot Use as a Violation of Self-Respect &
Journal of Applied Philosophy & \citep{van2025ai} \\
\hline
2025 & Friend, S., \& Goffin, K. &
Chatbot-fictionalism and empathetic AI: Should we worry about AI when AI worries about us? &
Philosophical Psychology & \citep{friend2025chatbot} \\
\hline
2025 & Reinecke, M. G., Ting, F., Savulescu, J., \& Singh, I. &
The Double-Edged Sword of Anthropomorphism in LLMs &
Proceedings & \citep{reinecke2025double} \\
\hline
2025 & Ciriello, R. F., Chen, A. Y., \& Rubinsztein, Z. A. &
Compassionate AI Design, Governance, and Use &
IEEE Transactions on Technology and Society & \citep{ciriello2025compassionate} \\
\hline
2025 & Nath, S. S. &
Simulated Souls: Investigating the Emotional Fallacy in Large Language Models &
SSRN & \citep{nath2025simulated} \\
\hline
2025 & Ferrario, A., Termine, A., \& Facchini, A. &
Social Misattributions in Conversations with Large Language Models &
SSRN & \citep{ferrario2025social} \\
\hline
2025 & Meier, R. &
Balancing Minds and Data: The Privacy Dilemma of LLMs and Anthropomorphism in LLMs &
Journal of Social Computing &  \citep{meier2025balancing} \\
\bottomrule
\end{tabularx}
\caption{Included records in the scoping review (publication status as of the final database search). 
At the time of the search, Ferrario et al.\ \citep{ferrario2025social} was available as a preprint on SSRN and is therefore listed under SSRN in the table; the paper has since been accepted for publication in the 
Proceedings of the 2025 AAAI/ACM Conference on AI, Ethics, and Society.}
\label{tab:included_records}
\end{table}

\subsection{Illustrative audit trail and compact codebook}
\label{app:audit_trail_codes}
We provide an illustrative audit trail showing how selected claims identified during charting were translated into provisional codes, higher-order themes, plain-language definitions, and linked research questions. The examples are illustrative and are drawn from a subset of the included studies.

\begin{table*}[h!]
\centering
\footnotesize
\renewcommand{\arraystretch}{1.12}
\setlength{\tabcolsep}{4pt}
\caption{Illustrative audit trail for the thematic synthesis.}
\label{tab:audit_trail_codes}
\begin{tabularx}{\textwidth}{
p{0.14\textwidth}
p{0.28\textwidth}
p{0.13\textwidth}
p{0.18\textwidth}
p{0.12\textwidth}
p{0.05\textwidth}}
\toprule
\textbf{Study} &
\textbf{Short extracted claims (paraphrased)} &
\textbf{Provisional code} &
\textbf{Plain-language definition} &
\textbf{Theme} &
\textbf{RQ} \\
\midrule

\citet{abercrombie-etal-2023-mirages} &
Linguistic factors (e.g., voice, prosody, and disfluencies) and content (e.g., responses to direct prompting) can elicit or steer users' anthropomorphic attributions. &
Human-like interaction features &
Observable design or interaction features that can make the system seem more human-like. &
Anthropomorphic cues &
RQ3 \\
\citet{shanahan2024talking} &
Terms such as ``belief,'' or ``understanding'' may be  convenient shorthands, but can mislead when they obscure that LLMs are fundamentally next-token predictors rather than grounded believers or knowers. &
Mental-state attribution; intentional stance &
Anthropomorphism is attributing human-like mental states to the LLM, often as a useful but  misleading explanatory shortcut. &
Cognitive/epistemic anthropomorphism; attribution account &
RQ1 \\
\citet{maeda2024human} &
Anthropomorphic chatbot features can act as social affordances that simulate reciprocal engagement, encouraging users to project stable social roles onto the system and thereby fostering parasocial trust. &
Parasocial role framing &
Users start to relate to the system as if it occupied a human social role within an asymmetric, one-sided relationship. &
Relational and social-role attributions &
RQ1/RQ2 \\
\citet{bakir2024deception} &
In conversational AI systems, emulated empathy can become anthropomorphic deception when signals of care or understanding invite users to treat them as they were genuine concern. &
Superficial state deception via emulated empathy &
The system displays cues of empathy or care that can be mistaken for genuine concern, raising ethical issues, creating misleading social expectations and dependence. &
Affect-based influence &
RQ2 \\
\citet{marchegiani2025anthropomorphism} &
``Anthropomorphic false beliefs'' can undermine autonomy when users misapply human-to-human norms on the mistaken assumption that a system understands, cares, or reasons like a human. &
Anthropomorphic false beliefs and user autonomy &
Users act on mistaken beliefs about human-like capacities of the system, so their decisions no longer reliably align with their own intentions. &
Anthropomorphic false beliefs; user autonomy &
RQ2 \\
\citet{teixeira2025emotional} &
In mental-health contexts, anthropomorphic AI should be deployed only as a supervised support tool, with transparency about AI identity, informed consent, crisis-detection and escalation paths, scope limitation, and governance measures. &
Supervised mental-health deployment safeguards &
Context-specific design and governance measures are proposed to preserve supportive interaction while preventing false therapeutic expectations. &
Design and communication controls &
RQ3 \\
\citet{gabriel2024ethics} &
Anthropomorphic AI assistant features should be assessed case by case through a risk-benefit analysis that considers user vulnerability, explicit AI-status disclosure, sandbox testing of cue-induced harms, and mitigation through redesign or participatory ``de-anthropomorphising'' interventions. &
Anthropomorphism risk mitigation &
Anthropomorphic features should be disclosed, tested, and, if harmful, mitigated in ways that protect vulnerable users and reduce risks such as persuasion, privacy loss, and emotional dependence. &
Design and communication controls; evaluation pipelines &
RQ3 \\
\citet{leis2025ethical} &
Mental-health chatbots (MHCBs) should be governed through an explicit ethical framework that combines bioethics and AI-ethics values to address anthropomorphism, deception, overreliance, and regulatory gaps. &
Value-based ethical framework for MHCBs &
Ethical analysis is translated into concrete design and deployment principles for vulnerable mental-health contexts. &
Ethical frameworks and toolkits &
RQ3 \\
\citet{ferrario2025social} &
Social misattributions are higher-order, contextualized forms of anthropomorphisation: through conversation and role-playing, users may negotiate social roles and qualities with conversational AIs that are not supported by their technical capabilities. &
Social misattributions &
Social-role attributions can emerge through conversational interaction and become problematic when users treat the system as capable of fulfilling roles it cannot genuinely sustain. &
Misplaced social-role attribution; mitigation interventions &
RQ2/RQ3 \\
\bottomrule
\end{tabularx}
\label{table:audit_trail}
\end{table*}

\end{document}